%% file: neurips_2021.tex
\documentclass{article}

     \PassOptionsToPackage{numbers, compress}{natbib}
\usepackage[final]{neurips_2021}

\usepackage[utf8]{inputenc} % allow utf-8 input
\usepackage[T1]{fontenc}    % use 8-bit T1 fonts
\usepackage{hyperref}       % hyperlinks
\usepackage{url}            % simple URL typesetting
\usepackage{booktabs}       % professional-quality tables
\usepackage{amsfonts}       % blackboard math symbols
\usepackage{nicefrac}       % compact symbols for 1/2, etc.
\usepackage{microtype}      % microtypography
\usepackage{xcolor}         % colors
\usepackage{graphicx}
\usepackage{subcaption}
\usepackage{amsmath}
\usepackage{amsthm}
\usepackage{wrapfig}
\usepackage{thmtools}
\usepackage{thm-restate}
\usepackage{cleveref}
\definecolor{Red}{rgb}{0.768, 0.054, 0.054}
\definecolor{Blue}{rgb}{0.152, 0.294, 0.925}
\definecolor{Green}{rgb}{0,0.4,0.7}
\hypersetup{
    colorlinks=true,
    citecolor=Red,
    linkcolor=Blue,
    urlcolor=Green,
    }
\usepackage{algorithm}
\usepackage{algpseudocode}
\usepackage[font=footnotesize]{caption}

\title{Mini-Batch Consistent Slot Set Encoder for\\ Scalable Set Encoding}

\author{Andreis Bruno$^{1}$, Jeffrey Ryan Willette$^{1}$, Juho Lee$^{1,2}$, Sung Ju Hwang$^{1,2}$\\
    KAIST $^{1}$, South Korea\\AITRICS $^{2}$, South Korea \\
    \texttt{\{andries, jwillette, juholee, sjhwang82\}@kaist.ac.kr}
}

\declaretheorem[name=Property,numberwithin=section]{property}
\declaretheorem[name=Proposition,numberwithin=section]{proposition}
\declaretheorem[name=Definition,numberwithin=section]{definition}

\begin{document}
%TODO
\maketitle
\input{sections/abstract}

\input{sections/introduction}
\input{sections/approach}
\input{sections/experiments}
\input{sections/related}
\input{sections/conclusion}

\begin{small}
\bibliographystyle{abbrvnat}
\bibliography{bibs/bibs}
\end{small}

\clearpage
\appendix
\section*{Appendix}
\input{sections/appendix}
\end{document}

%% file: sections/abstract.tex
\begin{abstract}\label{sec:abstract}
  Most existing set encoding algorithms operate under the implicit assumption that all the set elements
  are accessible, and that there are ample computational and memory resources to load the set into memory during training and inference. 
  However, both assumptions fail when the set is excessively large such that it is impossible to load all set elements 
  into memory, or when data arrives in a stream. To tackle such practical challenges in large-scale 
  set encoding, the general set-function constraints of permutation invariance and equivariance are 
  not sufficient. We introduce a new property termed Mini-Batch Consistency (MBC) that is required for 
  large scale mini-batch set encoding. Additionally, we present a scalable and efficient attention-based set encoding 
  mechanism that is amenable to mini-batch processing of sets, and capable of updating set representations 
  as data arrives. The proposed method adheres to the required symmetries of invariance and equivariance 
  as well as maintaining MBC for any partition of the input set. We perform extensive experiments
  and show that our method is computationally efficient and results in rich set encoding representations for
  set-structured data. %We provide source code at the following anomymized repository: \url{https://github.com/AnonymousRepo1111/SSE }.
\end{abstract}

%% file: sections/introduction.tex
\section{Introduction}
Recent interest in neural network architectures that operate on sets~\citep{deepsets,settransformer} has garnered momentum given that many problems in
machine learning can be reformulated as learning functions on sets. Problems such as point cloud classification \citep{modelnet}, image reconstruction
\citep{cnp,anp,celeba}, classification, set prediction \citep{slotattention}, and set extension can all be cast in this framework of learning
functions over sets. Given that sets have no explicit structure on the set elements, such functions are required to conform to symmetric properties
such as permutation invariance or equivariance for consistent processing.

\par A defining property of many practical functions over sets involves an \textit{encoding} of the input set to a vector representation, the
\textit{set encoding}. This set encoding is then used for downstream tasks such as reconstruction or classification. In DeepSets \citep{deepsets}, a
sum-decomposable family of functions is derived for a class of neural network architectures that encode a given set to such a representation.
However, the simplicity of the functions derived in DeepSets makes it ineffective for modeling pairwise interactions between the elements of the sets.
Set Transformers \citep{settransformer} remedy this by using Transformers \citep{transformer} to model higher order interactions resulting in rich and expressive
set representations. 
\input{figures/fullmodel/full_model}

\par In all these works, there is an implicit assumption in the experimental setup that the cardinality of the set is manageable, and that there are ample computational 
resources  available for processing all the elements during the set encoding process. However in real-world applications such as large scale point cloud
classification and data intensive applications in particle physics, the set size can be extremely large. In such cases, even if one has access to a set encoding
function that is linear w.r.t. the number of elements in the set, it is still impossible to encode such sets in a single batch since we may not be able to load the
full set into memory. Current encoding methods deal with this issue by sampling subsets of the full set (as a data preprocessing step) and encoding them to a representative
vector, as done in the ModelNet40~\citep{modelnet} experiments in \citep{deepsets, settransformer}. However, this results in the loss of information about the full set, 
which is undesirable when huge amounts of monetary resources are invested in obtaining data such as in the Large Hadron Collider (LHC) experiments, and hence is
imperative that the data is exploited for maximum performance on the given task. As such, it is necessary to have a set encoder 
that can take advantage of the full data, in a computationally efficient way.\input{figures/fullmodel/consistency_demonstration}
\vspace{-0.15in}
\par  To this end, when the set is too large to fit in the memory, or given as a stream of data, we propose to \emph{iteratively} encode it over multiple rounds, by randomly
partitioning it, encoding each random subset (mini-batch) with the set encoder at each round, and aggregating all the encodings. However, naive mini-batch
set encoding could compromise essential properties of the set encoder such as permutation invariance and equivariance. To use a concrete example, in
Figure~\ref{fig:consistency_demonstration} we show the performance of Set Transformer (in Negative Log-Likelihood) on an image reconstruction task. 
By processing the full set (in this case concatenations of pixel and coordinate values), the model performs as expected (\textit{SetTransformer} in
Figure~\ref{fig:consistency_demonstration}). However when we instead partition the 
set elements into mini-batches, independently encode each batch and aggregate them to obtain a single set encoding, we observe performance degradation in the model as shown 
in Figure~\ref{fig:consistency_demonstration} (\textit{SetTransformer(MiniBatch)} in Figure~\ref{fig:consistency_demonstration}). This degradation stems from the attention module 
in Set Transformer where mini-batch encoding and aggregation results in mini-batch inconsistent set encoding. 

In this work, we identify and formalize a key property, \emph{Mini-Batch Consistency} (MBC), that is necessary for mini-batch encoding of sets. 
In what follows, we present the necessary constraints required for MBC and propose a \textit{slot} based attentive set encoder, the \emph{Slot Set Encoder} (SSE) which is MBC.
SSE largely alleviates the performance degradation in mini-batch encoding of large sets. We depict MBC and non-MBC set encoders in Figure~\ref{eqn:minibatchencoding}.

Our contributions in this work are as follows:

\begin{itemize}
    \item {We identify and formalize a key property of set encoding functions called Mini-Batch Consistency (MBC), which requires that mini-batch encoding of a 
set is provably equal to the encoding of the full set (in Section~\ref{sec:minibatch}).}
    \item {We present the first attention based set encoder that satisfies MBC. Our method can efficiently scale to arbitrary set sizes by removing
    the dependence of self-attention in the transformer through the use of \textit{slots} which frees the transformer
from requiring to process the full set in a single pass (in Section~\ref{sec:slotencoder}).}
    \item {We perform extensive experiments on various tasks such as image reconstruction, point cloud classification and dataset encoding where we demonstrate that SSE
    significantly outperforms the relevant baselines (Section~\ref{sec:experiments}).}
\end{itemize}

%% file: figures/fullmodel/full_model.tex
\begin{figure*}
    \centering
    \includegraphics[width=\textwidth]{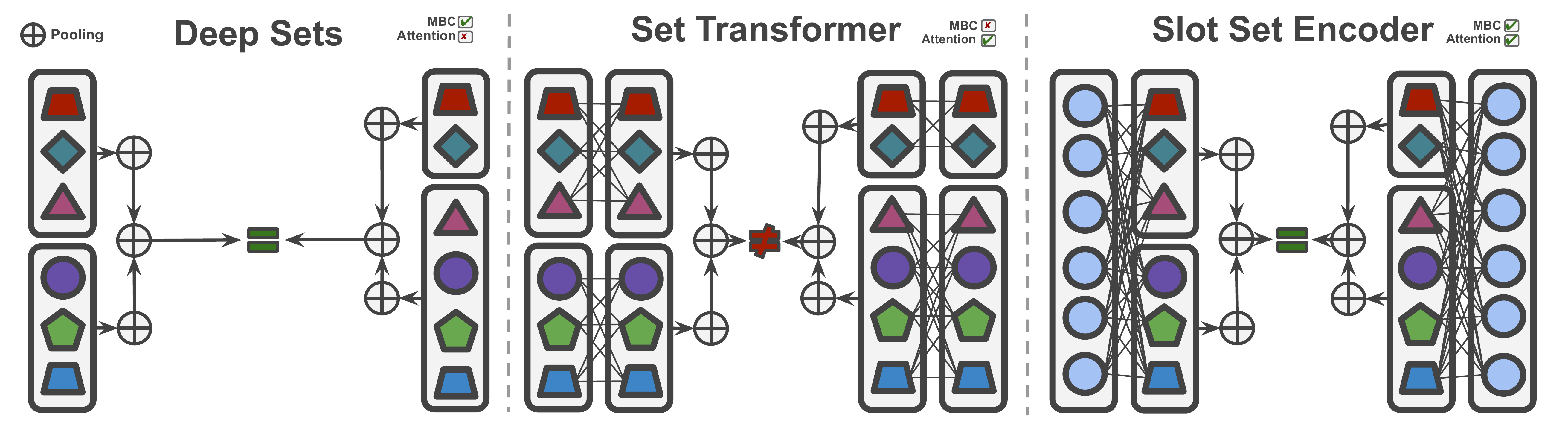}
    \caption{Mini-Batch Consistent Set Encoding. Each figure shows the processing of \textbf{(left)} mini-batch encoding of $\{s_1, s_2 \}\in S$, and \textbf{(right)} $\{s_1', s_2'\} \in S$. \textbf{Deep Sets:} can be made to encode batches consistently but cannot model pairwise interactions. \textbf{Set Transformer:} self-attention prevents mini-batch consistent set encoding. \textbf{Slot Set Encoder:} Attention w.r.t. parameterized slots allows for both attention and mini-batch consistency.}
    \label{fig:minibatchsetencoding}
    \vspace{-0.25in}
\end{figure*}

%% file: figures/fullmodel/consistency_demonstration.tex
\begin{wrapfigure}{r}{0.25\textwidth}
  \vspace{-0.15in}
  \begin{center}
    \includegraphics[width=0.25\textwidth]{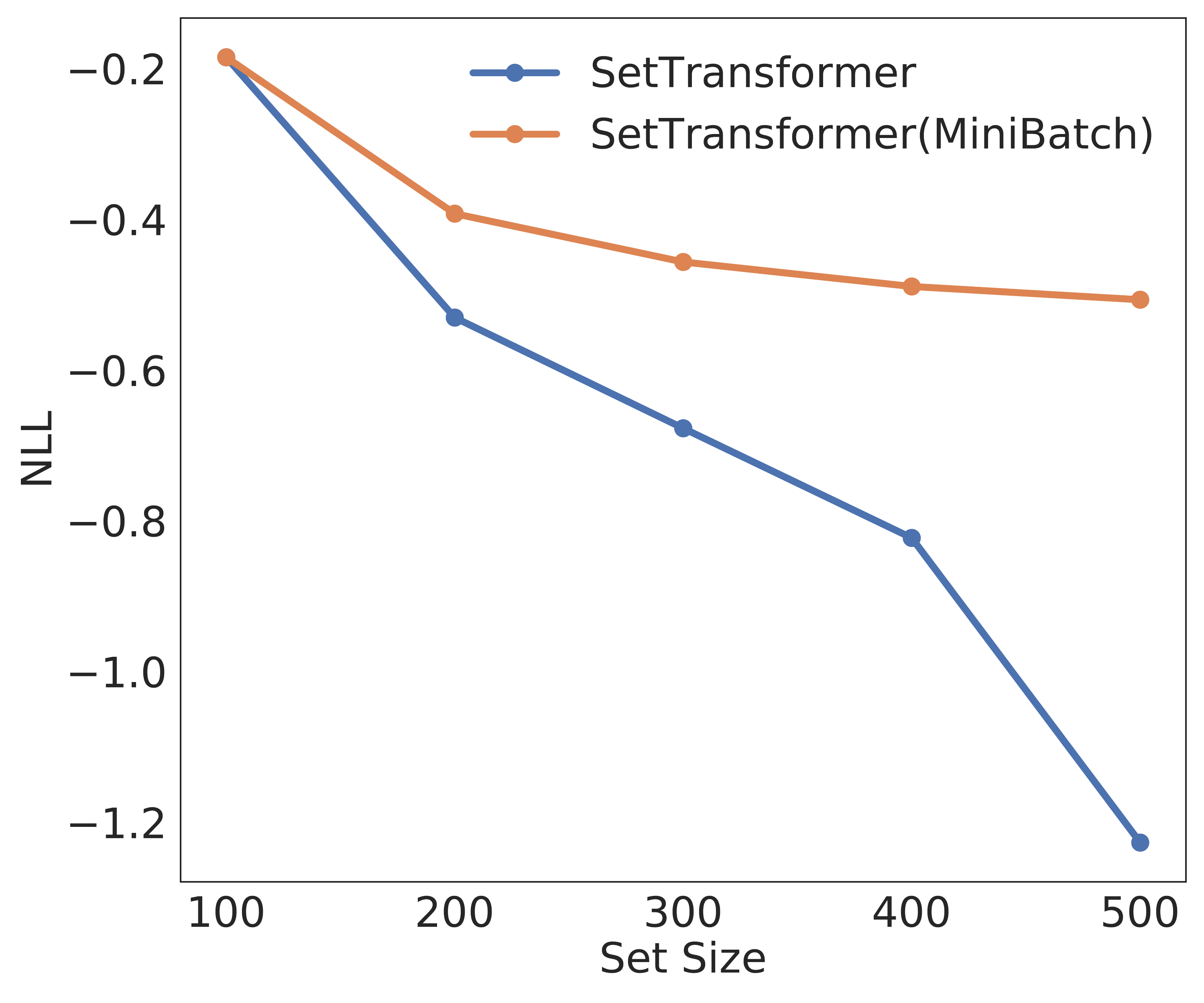}
  \end{center}
  \vspace{-0.1in}
  \caption{\footnotesize Performance degradation (CelebA image reconstruction) when evaluating Set Transformer \citep{settransformer} in a minibatch vs. non-minibatch setting. \label{fig:consistency_demonstration}}
  \vspace{-0.1in}
\end{wrapfigure}

%% file: sections/approach.tex
\section{Mini-Batch Consistent Set Encoding with Slot Set Encoders}\label{sec:approach}
\subsection{Preliminaries}\label{sec:preliminaries}
Suppose we are given a dataset of \textit{sets} $\mathcal{D} = \{X_i, \ldots, X_N\}$ where $|\mathcal{D}|=N$ and each $X_i \in \mathbb{R}^{n_i \times d}$
where $n_i$ is the number of elements in $X_i$ and each element $x_i^{(j)}$ (that is, the $j$th element of $X_i$) is represented by a $d$
dimensional tensor. For each $X_i$, the ordering of the $n_i$ individual elements is considered arbitrary and could be any permutation of the indices. We further assume
that both $n_i$ and $d$ can be large enough such that processing an instance $X_i$ is prohibitive both in terms of memory and computation.
As an example, the processing task could be encoding each element $X_i$ to a representative vector to be used for downstream tasks such as
set reconstruction, set classification or set prediction. Current set encoding methods such as \citet{deepsets} and \citet{settransformer} deal with such large sets 
by sampling a smaller subset of the original set as a preprocessing step. In essence, they approximate the full set with a random subset which can be undesirable when
we want to exploit all the data at our disposal for maximum performance.  In this section, we present
a method for encoding arbitrarily sized set data via a mini-batch encoding scheme. Our method can then \textit{iteratively} encode subsets of $X_i$ and aggregate 
the subset-encodings to obtain the full set representation. Our method is \textit{Mini-Batch Consistent}, invariant to the order of the subsets, 
and invariant to permutations on the set elements. For clarity, when we say a set $X_i$ is mini-batch processed, we mean that a random partition function can be applied to 
$X_i$ to obtain proper subsets of $X_i$. The subsets, which we call mini-batches, are then processed independently and aggregated to obtain a representation
for $X_i$.

\subsection{Mini-Batch Consistent Set Encoding}\label{sec:minibatch}
We consider a single element $X_i \in \mathcal{D}$ with $n_i$ elements each with $d$ dimensions. We truncate notation and write $X_i$, $n_i$
as $X$ and $n$ respectively since we present our algorithm for a single sample here. Our ultimate goal is to encode the set $X$ to a representation $Z
\in \mathbb{R}^{d^\prime}$ where $d^\prime$ is the dimension of the encoding vector. We assume that both $d$ and $n$ are large such that it is not
feasible to process $X$ as a whole. To get around this computational problem, we propose to perform mini-batch encoding of $X$
where the mini-batch samples are with respect to the number of elements $n$. Specifically, we partition $X$ such that $X$ can be written as a finite
union $X = X_1 \cup X_2 \cup \ldots \cup X_p$ and each $X_i$ in this partition has $n_i = |X_i|$ elements and $p$ is the total number of partitions.
We further assume that $X$ is partitioned such that we can efficiently process a given $\textit{mini-batch}$ $X_i$ in the partition. Each partition $X_i$ 
is encoded with a set encoding function $f(X_i)$ to obtain $Z_i$. We then define the set encoding problem as follows:
\begin{equation}\label{eqn:minibatchencoding}
  Z = g\big(f(X_1), f(X_2), \ldots, f(X_p) \big)
\end{equation}
where $g(\cdot)$ is an $\textit{aggregation}$ function that takes individual $Z_i$'s and combines them to obtain $Z$. We further require that the
functions $f$ and $g$ are \textit{Mini-Batch Consistent}.

\begin{property}[Mini-Batch Consistency]\label{prop:consistency}
  Let $X \in \mathbb{R}^{n \times d}$ be partitioned such that $X = X_1 \cup X_2 \cup \ldots \cup X_p$ and $f:\mathbb{R}^{n_i \times d} \mapsto
  \mathbb{R}^{d^\prime}$ be a set encoding function such that $f(X) = Z$. Given an aggregation function $g:\{Z_j \in \mathbb{R}^{d^\prime}\}_{j=1}^{p} \mapsto
  \mathbb{R}^{d^\prime}$, $g$ and $f$ are Mini-Batch Consistent if and only if
  \[g\big(f(X_1), \ldots, f(X_p)\big) = f(X)\]
\end{property}

Property \ref{prop:consistency} ensures that no information is lost by encoding the full set in independent mini-batches, and that the aggregation function $g$ 
guarantees the same output as encoding the full set in a single batch. Additionally, Property \ref{prop:consistency} is general and places no restriction on the form 
of the functions $f$ and $g$ except that when used together, MBC must be satisfied. In what follows, we define an attention-based set encoding function $f$ together with
aggregation functions $g$ that satisfy Property \ref{prop:consistency}.

\subsection{Slot Set Encoder}\label{sec:slotencoder}
In this section, we provide a formulation for an attention based set encoding function $f$ in Equation \ref{eqn:minibatchencoding} which utilizes
 slots \citep{slotattention}. Given an input set $X \in \mathbb{R}^{n \times d}$, we begin by sampling a random initialization for $K$ slots 
 $S \in \mathbb{R}^{K \times h}$ where $h$ is the dimension of each slot. Specifically,
\begin{equation}\label{eqn:slotencoder1}
  S \sim \mathcal{N}(\mu, \text{diag}(\sigma)) \in \mathbb{R}^{K \times h},
\end{equation}
where $\mu \in \mathbb{R}^{1 \times h}$ and $\sigma \in \mathbb{R}^{1 \times h}$ are learnable parameters. Optionally, instead of sampling random initialization
for the $K$ slots, we can instead designate deterministic learnable parameters 
$S_{\chi} \in \mathbb{R} ^{K \times h}$. An advantage of random initialization over learnable parameters is that at test time, we can increase or decrease $K$
for the randomly initialized model. In the ablation studies in section \ref{sec:ablation}, we explore the performance gains that result from the 
choice of slot initialization. The key design choice in the slot set encoder is that unlike attention based set encoding methods like \citet{settransformer},
we compute attention over $S$ (no self-attention) instead of computing it over the $n$ elements of $X$ (self-attention). This allows us to remove the dependence of 
the set encoding function on $n$, alleviating the requirement of processing the entire set at once, and allowing for MBC encoding of partitions of $X$ since 
the \textit{same} slots are used to encode \textit{all} mini-batches. Specifically, we compute dot product attention between $S$ and $X$ according to the following formulation:
\begin{equation}\label{eqn:slotencoder2}
\small 
  \text{attn}_{i,j} := \sigma(M_{i,j}) \hspace{0.3em} \text{where} \hspace{0.3em} M := \frac{1}{\sqrt{\hat{d}}}k(X) \cdot q(S)^{T} \in \mathbb{R}^{n \times K}
\end{equation}
where $k$ and $q$ are linear projections of $X$ and $S$ to a common dimension $\hat{d}$ and $\sigma$ is the sigmoid activation function. There is a key difference in our
attention formulation in that we use a sigmoid instead of a softmax to construct the attention matrix. Using a softmax, even when computed over S, breaks MBC as it requires
a batch dependent normalization term which will vary across different partitions of $X$. By using the sigmoid function, we are able to free our model of this
constraint and satisfy MBC. Finally, we weight the inputs based on the computed attention matrix to obtain the final slot set encoding $\hat{S}$ for $X$:
\begin{equation}\label{eqn:slotencoder3}
  \hat{S} := W^{T} \cdot v(X) \in \mathbb{R}^{K \times \hat{d}} \hspace{0.3em} \text{where} \hspace{0.3em} W_{i,j} := \frac{\text{attn}_{i,j}}{\sum_{l=1}^{K} \text{attn}_{i,l}}
\end{equation}
where $v$ is also a linear projection applied to $X$ and $W \in \mathbb{R}^{n \times K}$ are the weights computed over slots instead of elements. The normalization 
constant in $W$ occurs over the slot dimension $K$, and only contains a linear combination of a single $n_i \in X$ and slots $S$ and therefore has no dependence on 
other elements within the partition $X_i$. The Slot Set Encoder is fully described in Algorithm \ref{alg:slotsetencoder}.

\par
The Slot Set Encoder in Algorithm \ref{alg:slotsetencoder} is functionally composable over partitions of the input set $X$. More concretely, for \textit{a given}
slot initialization and \textit{any} partition of $X$,
\begin{equation}\label{eqn:slotencoder4}
  f(X) = g(f(X_1), f(X_2), \ldots, f(X_p))
\end{equation}
Where the aggregation function $g$ is chosen to satisfy Property \ref{prop:consistency}. This is convenient since it allows us to define $g$ from the following set of operators: $g \in \{\verb|mean, sum, max, min|\}$. We note that to satisfy MBC, we have chosen $g$ to be an associative function but it is entirely possible
to formulate a non-associative $g$ and corresponding $f$ that satisfies MBC and we leave that for future work. 

\input{algorithms/slotsetencoder}

\begin{proposition}\label{props:consistency}
  For a given input set $X \in \mathbb{R}^{n \times d}$ and slot initialization $S \in \mathbb{R}^{K \times d}$, the functions $f$ and $g$ as defined
  in Algorithm \ref{alg:slotsetencoder} are Mini-Batch Consistent for any partition of $X$ and hence satisfy Property \ref{prop:consistency}.
\end{proposition}

\begin{proposition}\label{props:perinvariance}
  Let $X \in \mathbb{R}^{n \times d}$ and $S \in \mathbb{R}^{K \times d}$ be an input set and slot initialization respectively.
  Additionally, let $\verb|SSE|(X, S)$ be the output of Algorithm \ref{alg:slotsetencoder}, and $\pi_X \in \mathbb{R}^{n \times n}$ and
  $\pi_S \in \mathbb{R}^{K \times K}$ be arbitrary permutation matrices. Then,
  \[\verb|SSE|(\pi_X \cdot X, \pi_S \cdot S) = \pi_S \cdot \verb|SSE|(X,S)\]
\end{proposition}

Proofs of Proposition \ref{prop:consistency} \& \ref{props:perinvariance} involve showing that each component in Algorithm \ref{alg:slotsetencoder} are both MBC, and permutation invariant or equivariant. Details can be found in the Appendix with discussions on why the baselines fail or satisfy MBC. 
%together with source code that
%empirically verifies Proposition \ref{prop:consistency} \& \ref{props:perinvariance} and discussions on why 
%baselines fail or satisfy MBC.

\subsection{Hierarchical Slot Set Encoder}\label{sec:hslotsetencoder}
The Slot Set Encoder can encode any given set $X \in \mathbb{R}^{n \times d}$ to a $\hat{d}$ dimensional vector representation. In the mini-batch
setting, all partitions are \textit{independently} encoded and aggregated using the aggregation function $g$. In many practical applications, it is
useful to model pairwise interactions between the elements in the given set since not all elements contribute equally to the set representation.
Indeed this is the key observation of \citet{settransformer} which models such pairwise and higher order interactions between set elements and obtain 
significant performance gains over \citet{deepsets} which assumes that all elements contribute equally to the set encoding. In the 
Slot Set Encoder, we would like to be able to model such pairwise and higher order interactions. 
However to satisfy Mini-Batch Consistency, we cannot model interactions between set elements given that the Slot Set Encoder removes all
dependencies on $n$. Hence, we propose to model interactions among \textit{slots} via a stack of Slot Set Encoders. 
Specifically, instead of using a Slot Set Encoder with $K=1$, we stack $T$ Slot Set Encoders each with $K_i$ slots and set $K_T=1$ for the final 
encoder. Concretely, if we let $\verb|SSE|$ be an instance of the Slot Set Encoder defined in Algorithm \ref{alg:slotsetencoder}, then we can 
define a series of such functions, $\verb|SSE|_1, \ldots, \verb|SSE|_T$. Then for a set $X$, a composition of these Slot Set Encoders 
is a valid set encoding function that satisfies all the requirements outlined in Section \ref{sec:slotencoder}, such that:
\begin{equation}\label{eqn:hslotsetencoder}
    f(X) = \verb|SSE|_T(\ldots \verb|SSE|_{2}(\verb|SSE|_{1}(X)))
\end{equation}
This gives a \textit{hierarchy} of Slot Set Encoders capable of modeling pairwise and higher order interactions between slots. The 
Hierarchical Slot Set Encoder is analogous to stacking multiple Induced Set Attention blocks followed by a final Pooling MultiHead
Attention used in Set Transformer \citep{settransformer}. The major difference is that our model still remains MBC, a property violated by Set Transformer.

\subsection{Approximate Mini-Batch Training of Mini-Batch Consistent Set Encoders}\label{sec:appminibatch}
Set encoding mechanisms such as \citet{deepsets} and \citet{settransformer} require that gradient steps are
taken with respect to the full set. However, in the Mini-Batch Consistent setting described so far, this is not feasible given large set sizes or
constraints on computational resources. A solution to this problem is to train on partitions of sets sampled at each iteration of the optimization process. 
We verify this approach empirically and show that the Slot Set Encoder presented so far can be trained on partitions of sets and still generalize 
to the full set at test time. In these experiments (Figure \ref{fig:incremental_set_encoding}), we train set encoders on subsets of sets sampled at each optimization
iteration and perform inference on the full set. Specifically, at some iteration $t$, we sample a mini-batch of size $B \times \tilde{n} \times d$,
where $B$ is the batch size and $\tilde{n}$ is the cardinality of a partition of a set with  $n$ elements ($ \tilde{n} < n$). For this empirical analysis, we use the 
CelebA dataset~\citep{celeba} where the pixels in an image forms a set and the task is to reconstruct the image using the set encoding of a few pixels given as  
context points. 

%% file: algorithms/slotsetencoder.tex
\begin{wrapfigure}{R}{0.52\textwidth}
\begin{minipage}{0.52 \textwidth}
    \vspace{-0.8cm}
    \begin{algorithm}[H]
       \caption{Slot Set Encoder. \\ Partitioned input $X = \{X_1, \ldots, X_p \}$ \\ Initialized slots $S\in \mathbb{R}^{K \times h}$ \\ aggregation function $g$.}
       \label{alg:slotsetencoder}
      \begin{algorithmic}[1]
        \State \textbf{Input:} $X = \{X_1, X_2, \ldots, X_p\}$, $S \in \mathbb{R}^{K \times h}, g$
        \State \textbf{Output:} $\hat{S} \in \mathbb{R}^{K \times d}$
         
        \State \textbf{Initialize} $\hat{S}$ 
        \State $S = \verb|LayerNorm|(S)$
        \For{$i=1$ {\bfseries to} $p$}
          \State Compute $\verb|attn|_i(X_i, S)$ using Equation \ref{eqn:slotencoder2}
          \State Compute $\hat{S_i}(X_i, \verb|attn|_i)$ using Equation \ref{eqn:slotencoder3}
          \State $\hat{S} = g(\hat{S}, \hat{S_i})$
        \EndFor
        \State \textbf{return} $\hat{S}$
      \end{algorithmic}
    \end{algorithm}
    %\vspace{-1.5cm}
\end{minipage}
\end{wrapfigure}

%% file: sections/experiments.tex
\section{Experiments}\label{sec:experiments}
We evaluate our model on the ImageNet \citep{imagenet}, CelebA \citep{celeba}, MNIST \citep{mnist} and ModelNet40 \citep{modelnet} datasets. Details on these 
datasets can be found in the Appendix. In these experiments, all sets arrive in a streaming setting in a mini-batch fashion where we only get subsets (batches) 
of the full set at an instance. Once the mini-batch is encoded, we only have access to the set encoding vector (we do not keep the original set elements 
since it requires further storage) which must be updated as more batches of the set arrive. We provide further implementation details in the Appendix.

\subsection{Image Reconstruction}
In this task, we are given a set of pixels (each 3 dimensional) with their corresponding coordinate points (each 2 dimensional), termed context
values, and the task is to reconstruct the full image given the coordinates of the pixels we wish to predict using only the context points. The
context points are 5 dimensional vectors formed by concatenating the pixels and their corresponding coordinates. We use Conditional Neural Processes
\citep{cnp}, CNP, for this task.  CNP has a set encoding layer for compressing the context points into a single vector representation that is then
passed to a decoder for constructing the full image. Here, our goal is to demonstrate that the Slot Set Encoder is a proper set encoding function
capable of learning rich set representations.  We sample 200, 400, and 500 context points for images of size $32 \times 32$ and 1000 context points
for images of size $64 \times 64$. As baselines, we use the mean set encoding mechanism of \citet{deepsets} and the Pooling MultiHead Attention (PMA)
Blocks of \citet{settransformer}. We compare this with a variant of the Slot Set Encoder with random slot initialization. All the methods encode the
given set to a 64 dimensional representative vector and are trained for 200 epochs.  Additionally,  for fair comparison, we only use a single Slot Set
Encoder in our models as well as a single PMA block for Set Transformers.

The results for these experiments are presented in Figure \ref{fig:small_scale_reconstruction} where it can be seen that the Slot Set Encoder model
learns richer set representations resulting in lower negative log-likelihood when trained on full sets. Additionally in Figure \ref{fig:incremental_set_encoding} we
demonstrate the scalability of the Slot Set Encoder where models initially trained on 1000 context points (in Figure \ref{fig:small_scale_reconstruction}) are used to encode
larger sets of context points (2000 - 4000) at test time. Since Set Transformer is not Mini-Batch Consistent, it cannot make use of the additional context points as it violates Property
\ref{prop:consistency} hence in the constraints of the experiments, we use a 1000 context points for Set Transformer. DeepSets and Slot Set Encoders can utilize the additional data but
the representation obtained by the Slot Set Encoder is richer in its representation power resulting in significant performance gains at test time as well as better generalization to larger
set sizes than was used during training. In Section \ref{sec:ablation} we further discuss Figure \ref{fig:incremental_set_encoding} in the context of constrained resources and streaming data. 

\subsubsection{Ablation}\label{sec:ablation}
The Slot Set Encoder presented so far has various components and evaluation modes that can have an impact on the richness of the resulting set
representation. In this subsection, we perform extensive ablation studies to explore how these various components contribute to the performance of the
encoder. 
\input{figures/celeba/celeba_nll}
\par \textbf{Constrained Resources and Streaming Data}
In order to demonstrate the utility of the models that satisfy Property \ref{prop:consistency}, we simulate cases where
the computational resources are limited in terms of memory. Specifically, we assume that our computational budget allows us to only compute the
set representation for 1000 context points at a time. Under these constraints, the Set Transformer can only be used to encode 1000 context points
since the Softmax attention layer requires all the set elements to be in memory. For MBC DeepSets, and Slot Set Encoder, we can use mini-batches of
of size 1000 and iteratively update the set representation until the full set has been encoded. Additionally, this setting is akin to the case of streaming data where
the set elements arrive or are obtained at irregular intervals and hence the set encoding representation must be incrementally updated. 

We present the results of this experiment for an input image with 4096 pixels with 1000 context points randomly selected and encoded at each time step. From
Figure \ref{fig:incremental_set_encoding} it can be seen that for both DeepSets and the Slot Set Encoder, the more batches we encode, the better the
performance. However, the Slot Set Encoder significantly outperforms DeepSets due to it's ability to model interactions between the set elements and slots.
Additionally, it can be seen that the MBC models get better as more data arrives justifying our initial motivation for making use of all the available set elements 
instead of using a randomly sampled subset (in the preprocessing stage) in place of the full set.

\par \textbf{Mini-Batch Training of Mini-Batch Consistent Set Encoders}
In Section \ref{sec:appminibatch}, we proposed to train the Slot Set Encoder by approximating the full training procedure on partitions of the set. As
we highlighted, under memory constraint and large set size assumptions, we cannot take gradient steps with respect to the full set. To demonstrate
that this mini-batch training generalizes when the full set is processed at inference time, we train a model on 1000 elements and test it on larger set
sizes. This result is demonstrated in Figure \ref{fig:incremental_set_encoding} on the CelebA dataset where the model performance improves as larger sets are
processed. In Figure \ref{fig:generalization_to_small_sets}, we take the same model and evaluate it on smaller set sizes of 100-500 context points and 
compare it to both DeepSets and Set Transformers where again we find that the Slot Set Encoder perform significantly better. Additionally, Figure
\ref{fig:training_plot} shows that the models trained in this fashion can achieve a faster convergence rate. We find that
for the problems we consider in this work, this mini-batch training is sufficient and generalizes at inference time.
\input{figures/celeba/celeba_nll_2}
\par \textbf{Choice of Aggregation Function}
In Section \ref{sec:slotencoder}, we stated that we choose the aggregation function $g$ from the set $\{\verb|mean, sum, max, min|\}$.
On the CelebA experiments, we explore the effects of making such a choice for the model with random slot initialization. In Figure
\ref{fig:choice_of_g_random}, we take a model trained on sets with cardinality 1000 and evaluate it on larger sets at test time. As can be
inferred, the aggregation functions $\verb|mean|$ and $\verb|sum|$ consistently outperform the $\verb|min|$ and $\verb|max|$ for this task. This is
intuitive since in the image reconstruction task, an aggregation function that considers the contribution of all pixels is necessary. In Section
\ref{sec:pointcloudclassification}, we use a model with $\verb|max|$ as the aggregation function for Point Cloud Classification where it performs
better than the other options.  We find that the choice of the aggregation function $g$ is very much informed by the task at hand. We
observe a similar trend for the Slot Set Encoder with deterministic slot initialization and report this result in the Appendix.

\par \textbf{Deterministic (D) vs Random (R) Slot Initialization}
For slot $\textit{initialization}$, we have the choice of using deterministic or random (with learnable $\mu$ and $\sigma$) slots. In this choice, we
find in almost all our experiments that random initialization performs better across all the tasks we investigated. This trend can be observed in Figure
\ref{fig:random_vs_learned} where using random slot initialization results in lower negative log-likelihoods. To explain this behaviour, we reiterate
that $slots$ attend to different portions of the input set and hence the usage of random initialization encourages the model to attend to a wider
coverage of the input sets during training as opposed to deterministic slot initialization.

\par \textbf{Number of Slots}
When we define slots $S \in \mathbb{R}^{K \times h}$, we must decide the number of slots $K$ to use and we explore the effects of this choice on model
performance here. We fix a model that encodes an input set to a 64 dimensional vector and train multiple models with varying $K$ 
(specifically, we test for $K = 1, 32, 64, 128, 256$) and we observe that using very small or large number of slots can have a negative effect
on the performance of the model for both deterministic and random slot initialization. For this specific model, we find that selecting $K$ between 
32 and 128 performs best. Additionally, we find that this parameter is also very dependent on the task. More details are provided in the Appendix.

\par \textbf{Dimension of Slots}
Additionally, for slots $S \in \mathbb{R}^{K \times h}$, we can choose the dimension of each slot $h$ arbitrarily since the projection layer will
eventually reduce it to $\hat{h}$. Similarly to the experiments on the number of slots, we fix a model that encodes sets to 64 dimensions using a
single Slot Set Encoder and vary the dimensions of the slot keeping $K=1$. We experiment with $d= 32, 64, 128, 256$ and find that this parameter has a
significant impact on the performance of the model. Specifically,  we find that for slots with random initialization, the dimension of the slots can have
a negative impact if not  chosen properly. For deterministic initialization, increasing the slot dimension generally results in better
performance. In our experiments involving randomly initialized slots, we train multiple models with varying $h$ and pick the best one. The results 
discussed here can be found in the Appendix.

\par \textbf{Hierarchical Slot Set Encoder}
Finally, we explore stacking multiple layers of Slot Set Encoders based on the hierarchical formulation presented in Section
\ref{sec:hslotsetencoder}. We start with a single Slot Set Encoder with $K=1$ and $h=32$ and for each hierarchy, we double the slot dimension.  For
this experiment, we find that stacking multiple Slot Set Encoders \textit{for the tasks we consider} provides marginal performance gains and can
result in performance degradation when the hierarchy gets very deep. This is shown in Figure \ref{fig:slot_hierarchy} where the
performance of the model degrades after 4 hierarchies. 

\input{tables/modelnet/modelnet}
\input{figures/centroid_prediction/centroid_prediction_line_imagenet}
\subsection{Point Cloud Classification}\label{sec:pointcloudclassification}
Using the ModelNet40 dataset, we train models that encode an input set (point cloud) to a single set representation that is used to classify the
object as one of 40 classes. We follow the experimental settings of \citet{deepsets} and \citet{settransformer} where for point clouds
with cardinality less than 5000, all the compared methods encode the point cloud to a 256 dimensional vector and 512 for point clouds with 5000 points
or more. Set Transformer and DeepSets use a proprietary version of the ModelNet40 dataset however in our experiments, we use the raw ModelNet40 dataset
which can sample an arbitrary number of points, and sample points for each instance randomly at every iteration. 

The ModelNet40 dataset is difficult to classify when the number of elements in the set is relatively small. This is shown in Table
\ref{table:modelnet} where we test the two variants of our model with both deterministic (D) and random (R) slot initialization. The Slot Set Encoder
models outperform DeepSets and Set Transformer. As the number of points increase to 5000, DeepSets and the Slot Set Encoder model achieve similar
performance as expected. From the Table \ref{table:modelnet}, it can be seen that the two competing models are DeepSets and our models and hence we check the
generalization of both models in the Mini-Batch testing setting where a model trained on sets with 100 elements is tested on 200, 300, 400 and 500
elements. As can be seen from Figure \ref{fig:deepset_degradation}, although DeepSets is Mini-Batch Consistent, it experiences performance degradation when the set
size grows on the ModelNet40 dataset. Conversely, the Slot Set Encoder based model gets better as the set size increases at test time showing that
it generalizes better than DeepSets with Mini-Batch training (Section \ref{sec:appminibatch}). 

\subsection{Cluster Centroid Prediction}\label{sec:centroid_main}
In order to show the effectiveness of Mini-Batch Consistent set encoding on sets of high cardinality, we experiment with Prototypical Networks \citep{protonet} for classification on the ImageNet and MNIST datasets (see the Appendix for MNIST results). We use a pretrained ResNet50 backbone with the final layer removed, and train the set-encoding layers for
DeepSets/SSE in the latent feature space. We train for 20-way classification with 2 support and 2 query instances. Each support set is
encoded to a 128 dimensional vector (centroid), and we then train the model to minimize the Euclidean distance between the query instances and the corresponding
class centroid. There is no decoder in this experiment, as we only utilize the pooled representation for inference. We investigate the effects of testing on both lower and higher 'way' (5-1000 way) problems (Figure~\ref{fig:centroid_prediciton_line_imagenet}) and
different encoded support set sizes (Figure~\ref{fig:centroid_prediction_shots}). Given the large size of the set, we compare our model with DeepSets,
as it is the only other MBC set encoder. In Figure \ref{fig:centroid_prediction_shots} we plot accuracy vs. encoded set size for various 'ways'. In both the 1 and 20 way tasks, our SSE and Deepsets show similar performance, we attribute this to the fact that 1 way classification provides minimally informative singleton sets, and 20-way
classification is the setting in which the model was trained. The Slot Set Encoder generalizes better on the other 10 and 1000 way experiments which differ from the
training setting.
\input{figures/centroid_prediction/centroid_prediction_shots}

%% file: figures/celeba/celeba_nll.tex
\begin{figure}[t]
\begin{center}
	\begin{subfigure}{.25\linewidth}
		\includegraphics[width=\textwidth]{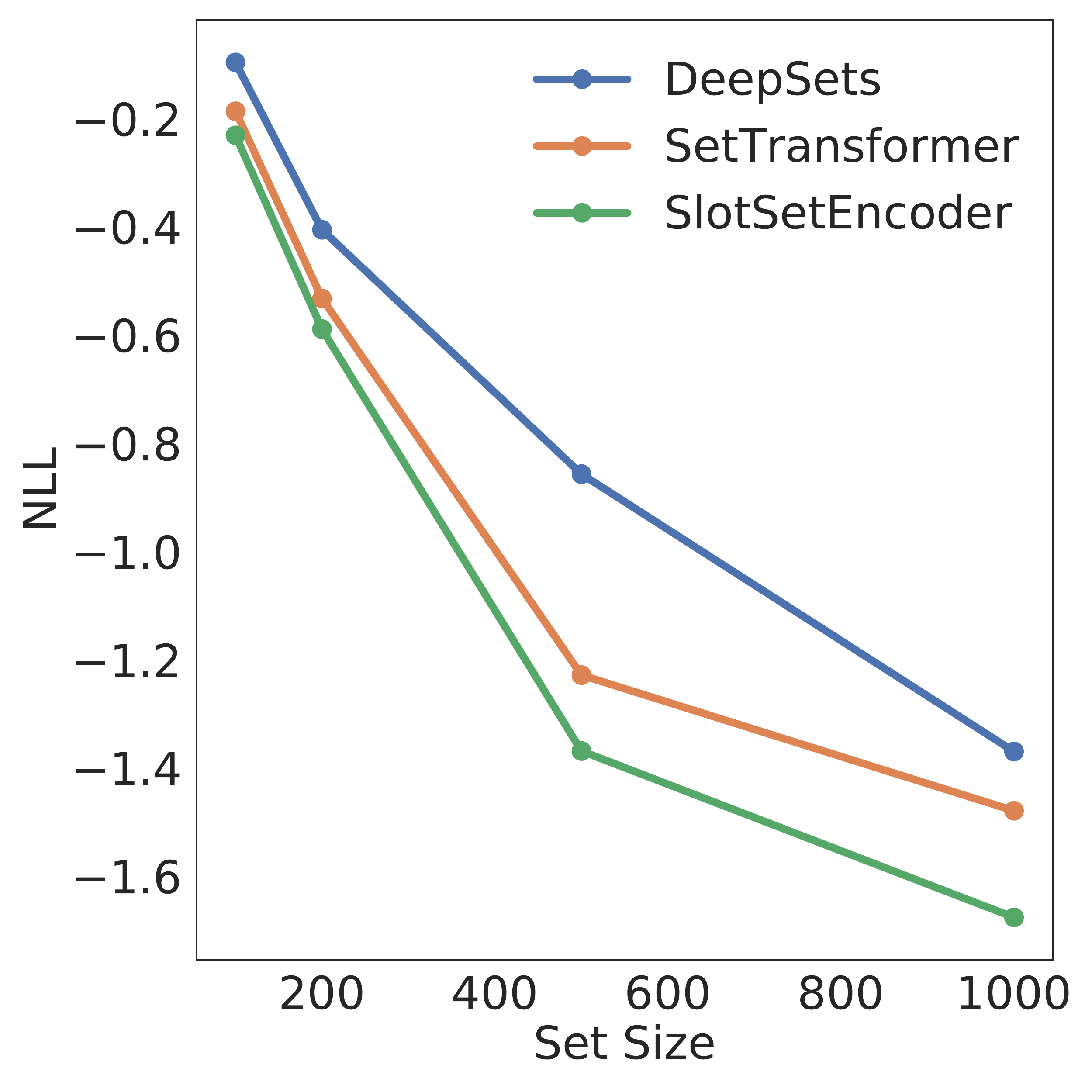}
		\caption{\small NLL on CelebA}
		\label{fig:small_scale_reconstruction}
	\end{subfigure}%
	\begin{subfigure}{.25\linewidth}
		\includegraphics[width=\textwidth]{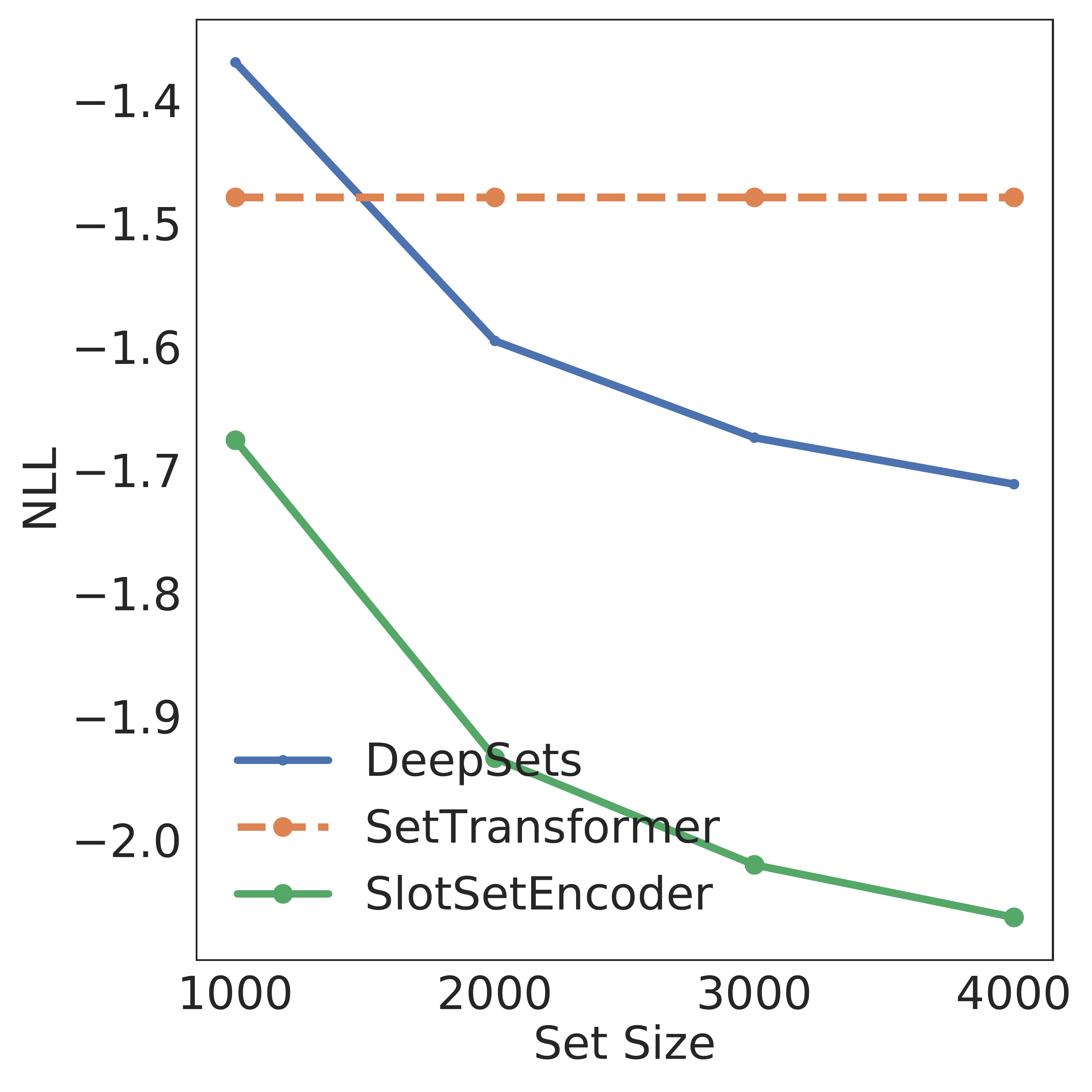}
		\caption{\small Mini-Batch Testing (L)}
		\label{fig:incremental_set_encoding}
	\end{subfigure}%
	\begin{subfigure}{.25\linewidth}
		\centering
		\includegraphics[width=\textwidth]{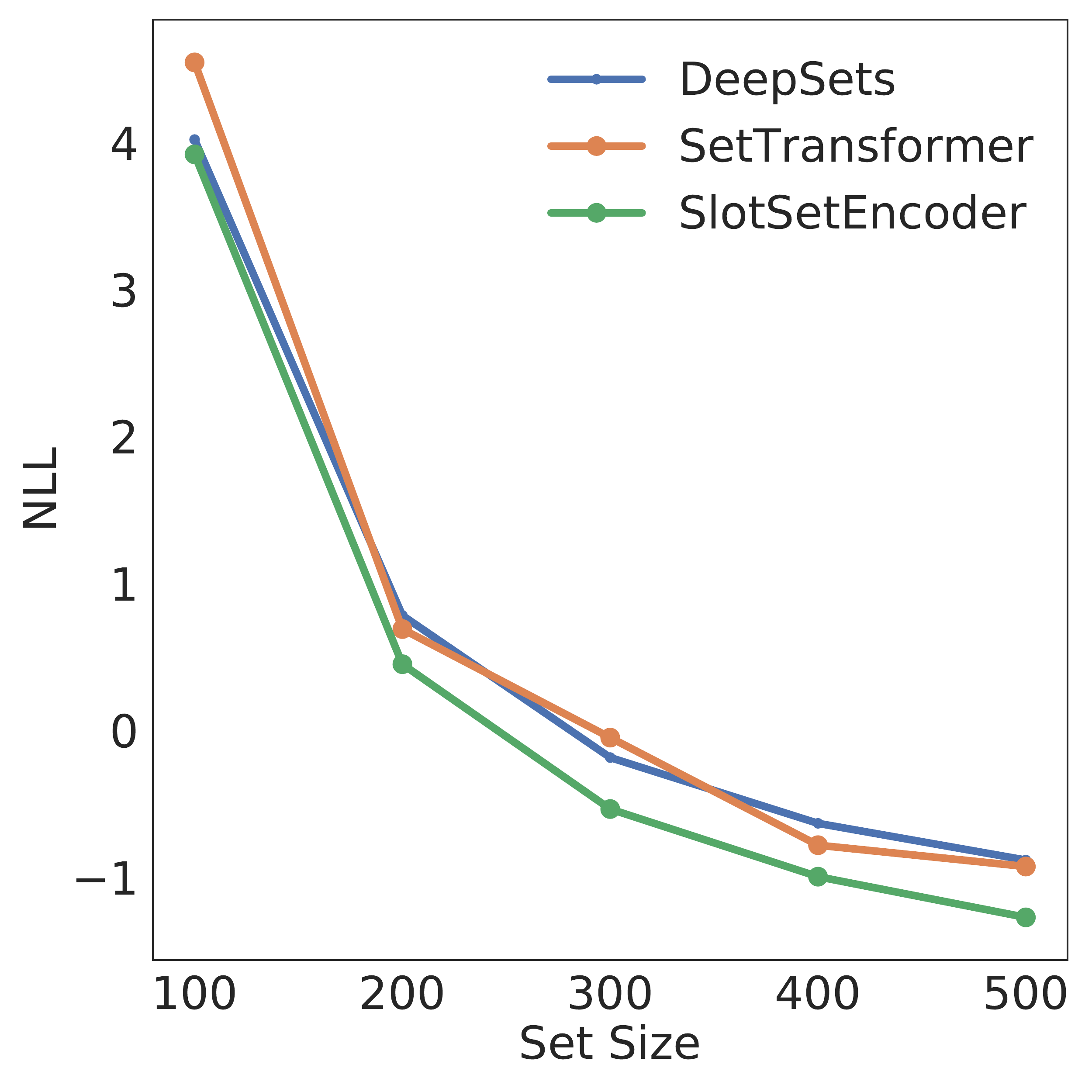}
		\caption{\small Mini-Batch Testing (S)}
		\label{fig:generalization_to_small_sets}
	\end{subfigure}%
	\begin{subfigure}{.25\linewidth}
		\centering
		\includegraphics[width=\textwidth]{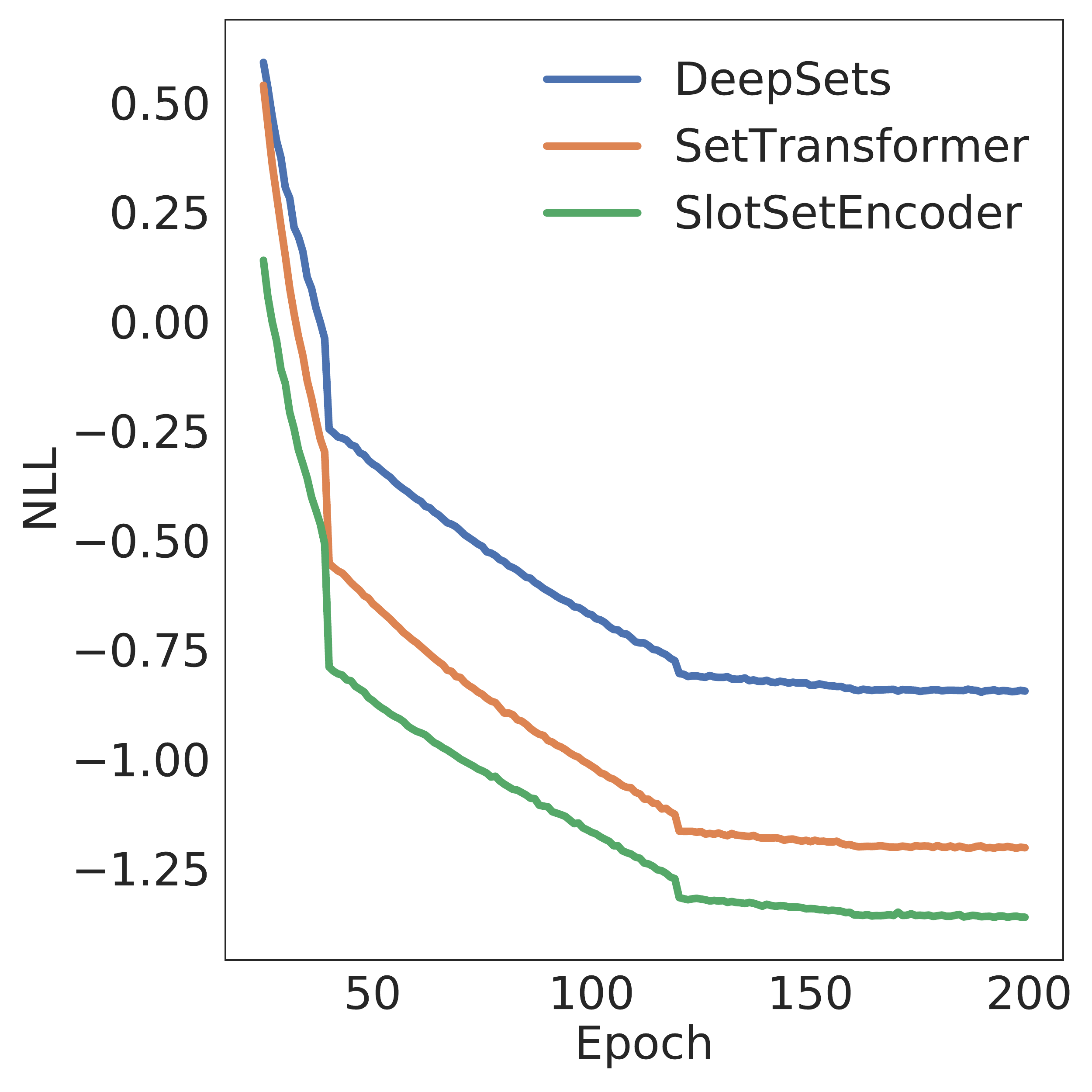}
		\caption{\small Convergence Plot}
		\label{fig:training_plot}
	\end{subfigure}
	
	\caption{\small 
	\textbf{(a)} NLL for different set sizes (full sets, no mini-batch processing is used here) on the image reconstruction task. 
	\textbf{(b)} Mini-Batch Testing of the reconstruction model (1000 set sized model in Figure \ref{fig:small_scale_reconstruction}) trained on 1000 pixels and evaluated on large (L) set sizes
	of 1000-4000 pixels processing mini-batches of size 1000 at a time.
	\textbf{(c)} Mini-Batch Testing of reconstruction model(same as in Figure \ref{fig:incremental_set_encoding}) on smaller (S) set sizes of 100-500 pixels processing mini-batches of size 100 at a time. 
	\textbf{(d)} Convergence plots for the considered Set Encoders on the image reconstruction task with the Slot Set Encoder trained according to Section \ref{sec:appminibatch}.
	} 
	\vspace{-0.15in}
\end{center}
\end{figure}

%% file: figures/celeba/celeba_nll_2.tex
\begin{figure}[t]
    \begin{center}
    \begin{subfigure}{.25\linewidth}
		\centering
		\includegraphics[width=\textwidth]{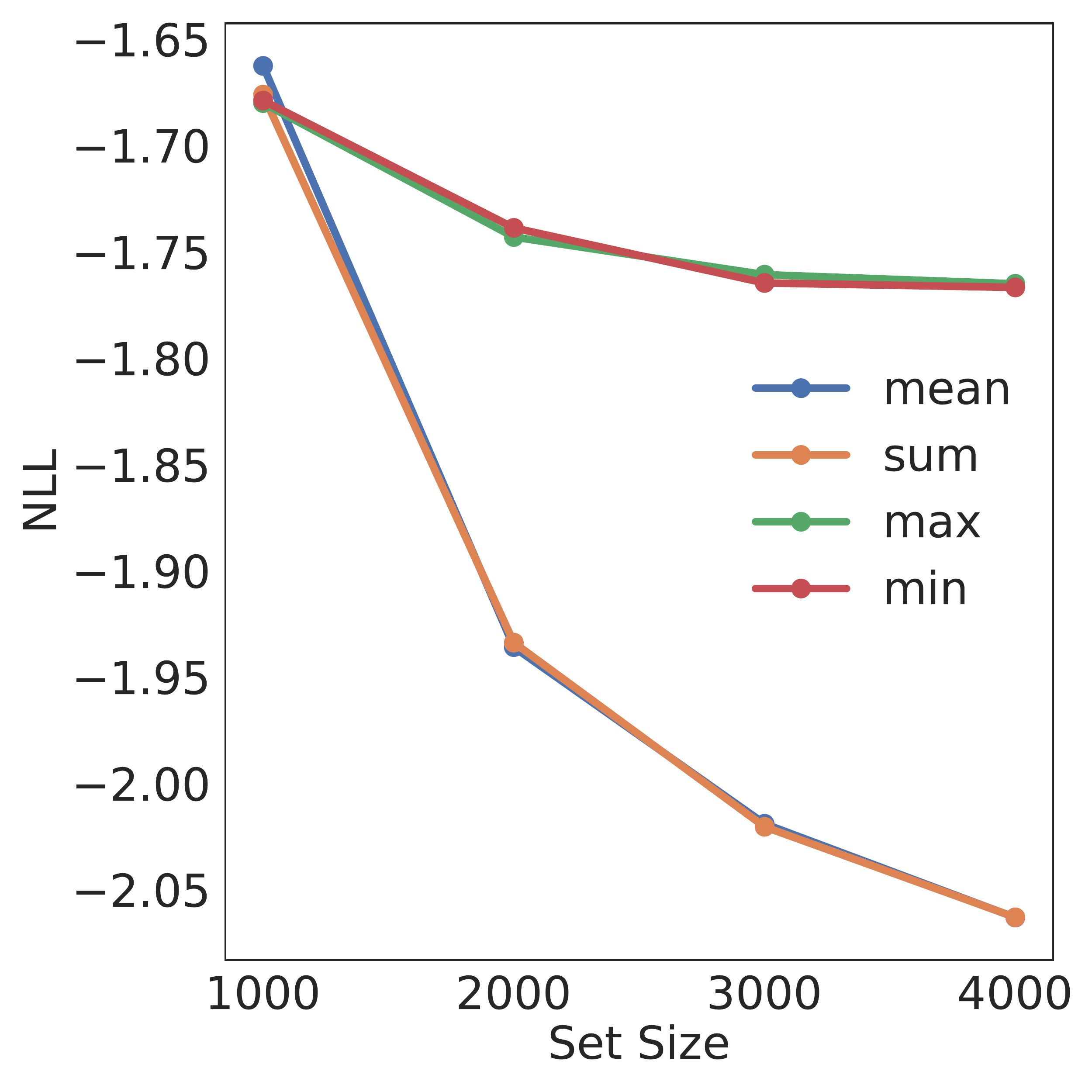}
		\caption{\small Choice of $g$ }
		\label{fig:choice_of_g_random}
	\end{subfigure}%
	\begin{subfigure}{.25\linewidth}
		\centering
		\includegraphics[width=\textwidth]{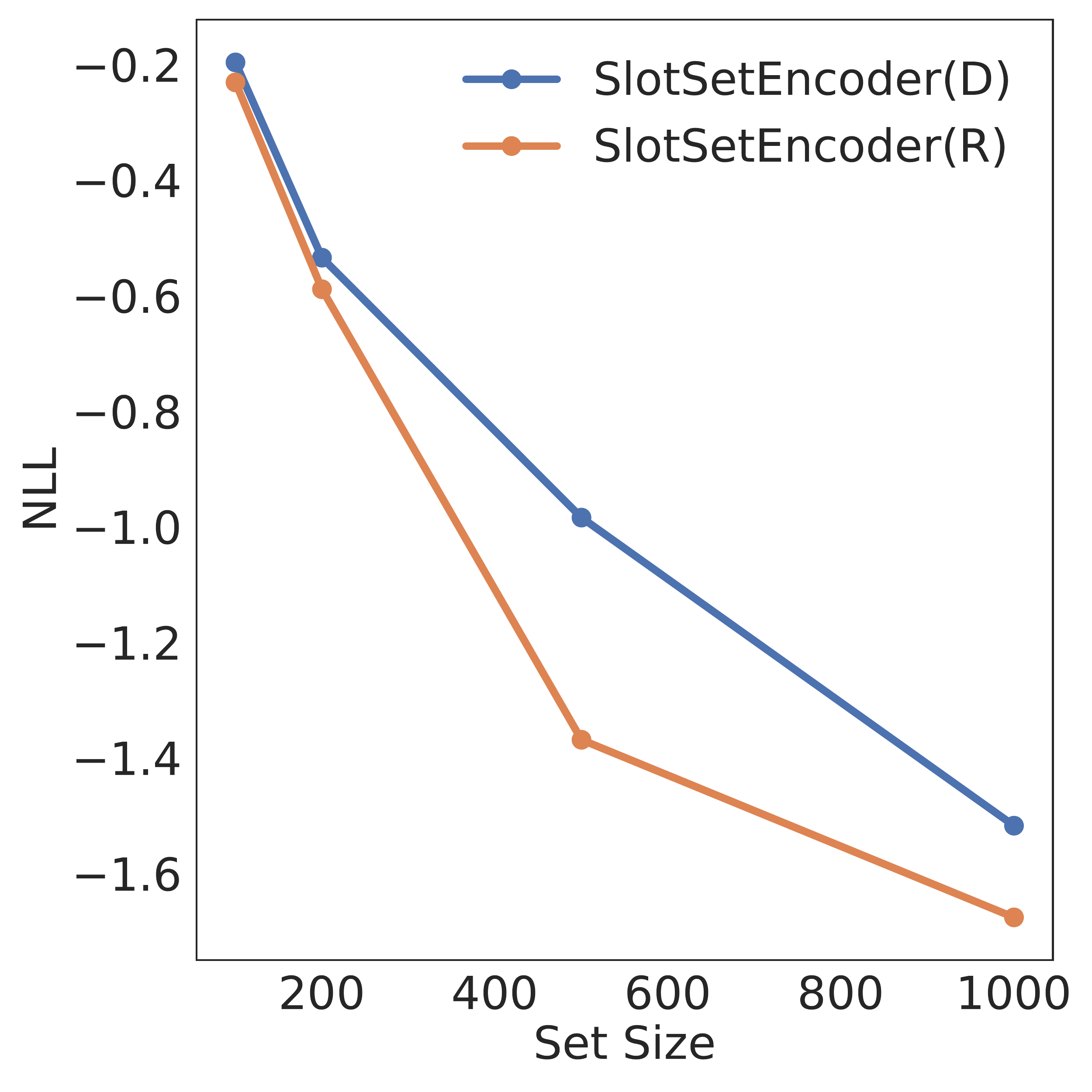}
		\caption{\small Slot Initialization}
		\label{fig:random_vs_learned}
	\end{subfigure}%
	\begin{subfigure}{.25\linewidth}
		\centering
		\includegraphics[width=\textwidth]{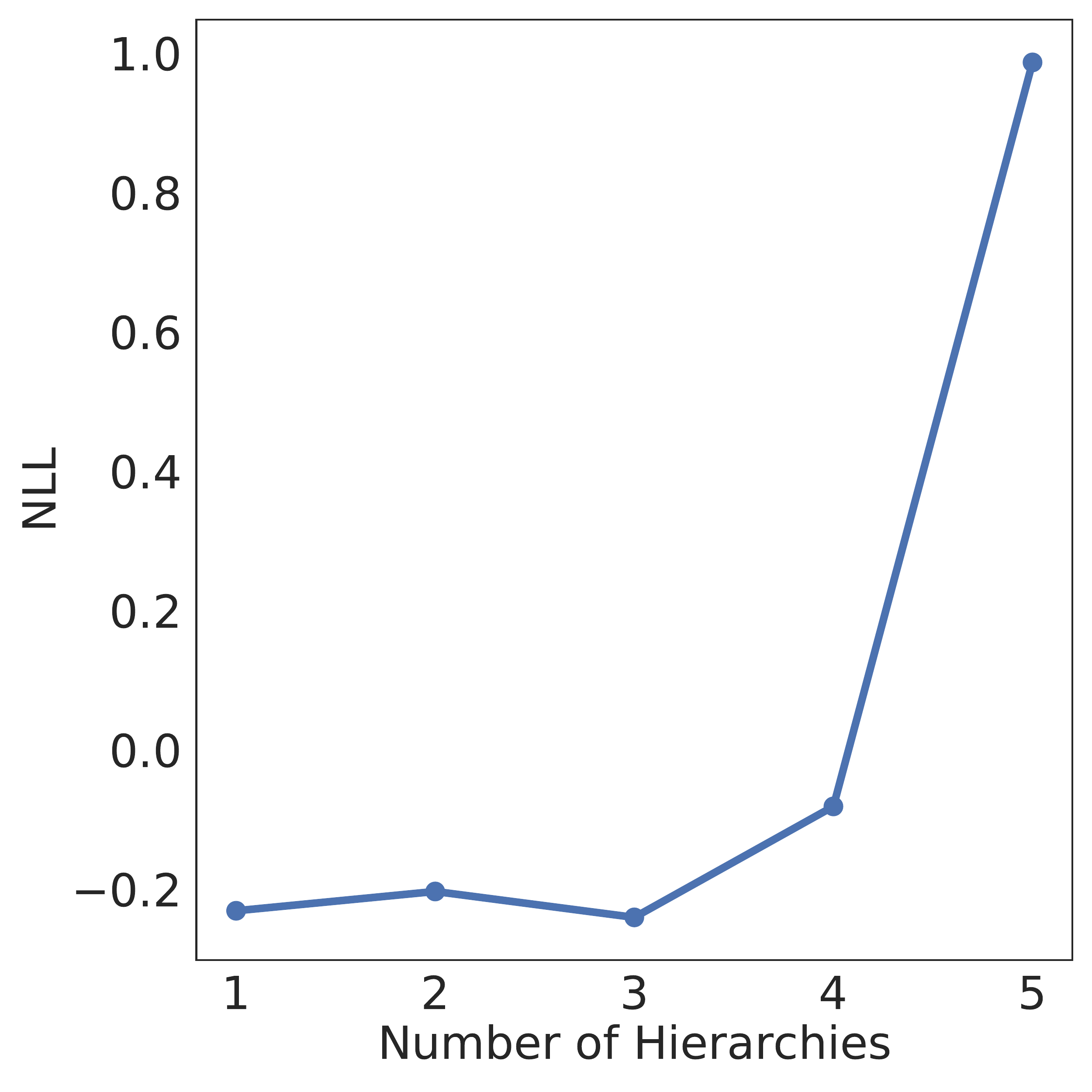}
		\caption{\small Hierarchical Slots}
		\label{fig:slot_hierarchy}
	\end{subfigure}%
	\begin{subfigure}{.25\linewidth}
		\centering
		\includegraphics[width=\textwidth]{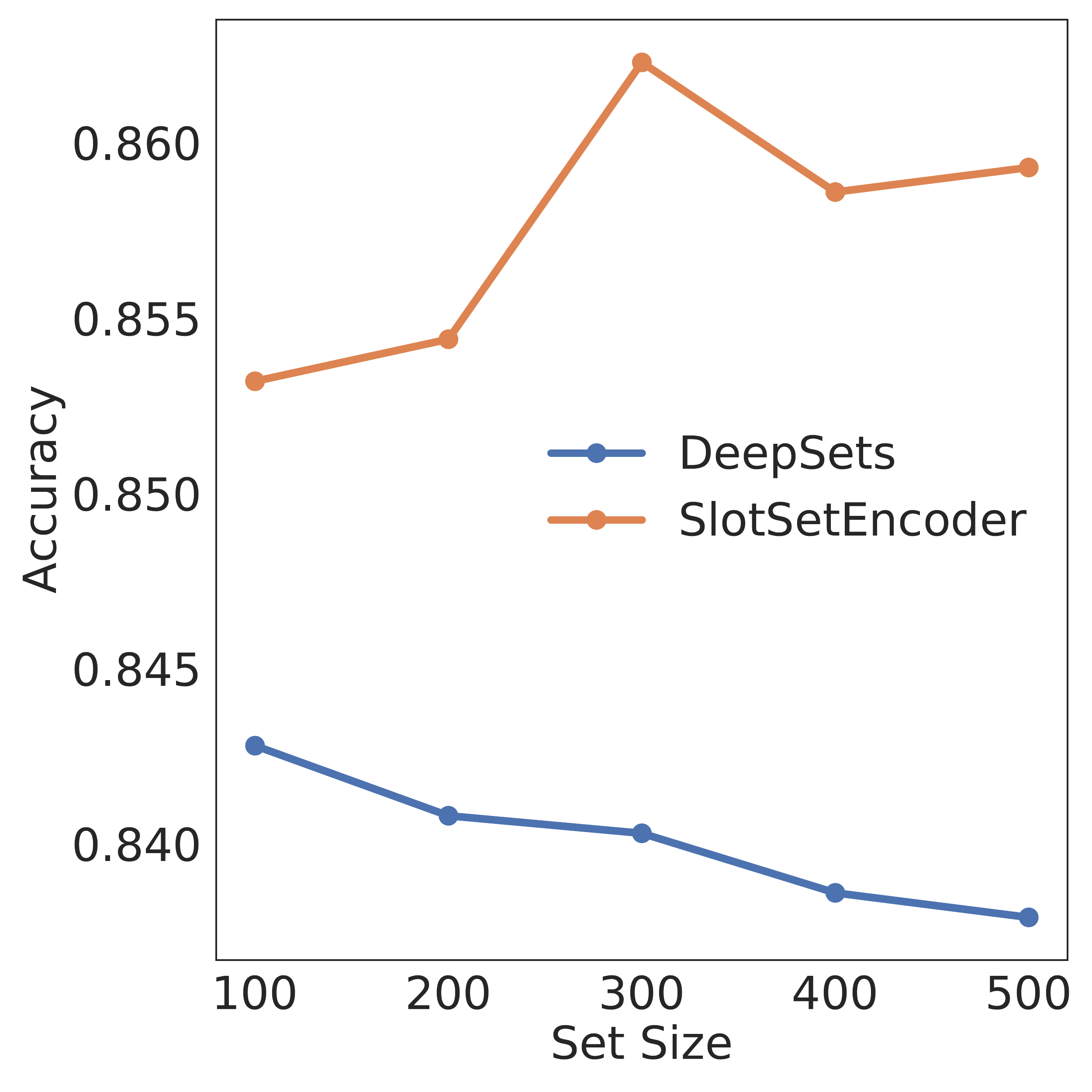}
		\caption{\small Mini-Batch Testing}
		\label{fig:deepset_degradation}
	\end{subfigure}
	\caption{\small 
	\textbf{(a)} Effects of the aggregation function $g$ on model performance. 
	\textbf{(b)} Effects of using random and deterministic slot initialization.
	\textbf{(c)} Effects of stacking multiple hierarchies of the Slot Set Encoder.
	\textbf{(d)} Accuracy of
	mini-batch testing of MBC models on the ModelNet40 dataset.}
    \end{center}
    \vspace{-0.30in}
\end{figure}

%% file: tables/modelnet/modelnet.tex
\begin{table*}
\begin{center}
  \caption{\small Test Accuracy for Point Cloud Classification with varying set sizes.\label{table:modelnet}}
  \resizebox{\linewidth}{!}{
  \begin{tabular}{lccccc}
  \toprule
  Model             &   100                 & 200                & 500                 & 1000                & 5000                 \\
  \midrule
  DS		    &   0.8428 $\pm$ 0.0027 & 0.8518 $\pm$ 0.0032 & 0.8574 $\pm$ 0.0017 & 0.8528 $\pm$ 0.0020 & \textbf{0.8883 $\pm$ 0.0008} \\
  ST		    &   0.8258 $\pm$ 0.0042 & 0.8443 $\pm$ 0.0023 & 0.8545 $\pm$ 0.0019 & 0.8511 $\pm$ 0.0018 & 0.8604 $\pm$ 0.0009 \\
  \midrule
  SSE(D)	    &   \textbf{0.8556 $\pm$ 0.0048} & 0.8633 $\pm$ 0.0038 & \textbf{0.8710 $\pm$ 0.0040} & 0.8720 $\pm$ 0.0009 & 0.8784 $\pm$ 0.0023 \\
  SSE(R)	    &   0.8532 $\pm$ 0.0009 & \textbf{0.8667 $\pm$ 0.0028} & 0.8699 $\pm$ 0.0020 & \textbf{0.8743 $\pm$ 0.0021} & 0.8853 $\pm$ 0.0012 \\
  \bottomrule
  \end{tabular}
  }
\end{center}
\vspace{-0.2in}
\end{table*}

%% file: figures/centroid_prediction/centroid_prediction_line_imagenet.tex
\begin{wrapfigure}{r}{0.25\textwidth}
    \vspace{-0.6in}
    \centering
    \includegraphics[width=0.25\textwidth]{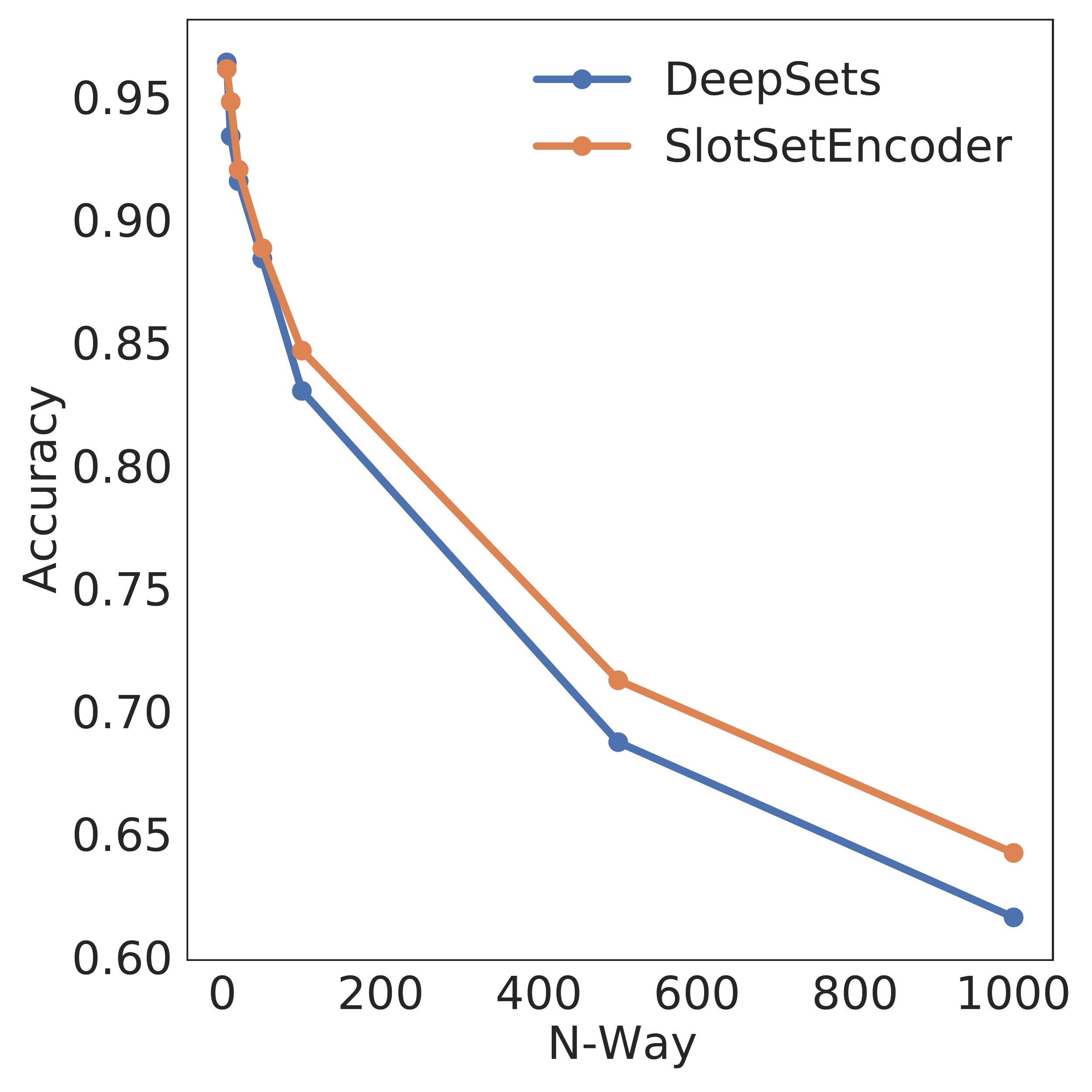}
    \caption{Accuracy vs. 'way' in the centroid prediction task. As the problem becomes harder, our Slot Set Encoder outperforms DeepSets.}
    \label{fig:centroid_prediciton_line_imagenet}
    \vspace{-0.2in}
\end{wrapfigure}

%% file: figures/centroid_prediction/centroid_prediction_shots.tex
\begin{figure*}
    \begin{center}
    \begin{subfigure}{.25\linewidth}
		\centering
		\includegraphics[width=\textwidth]{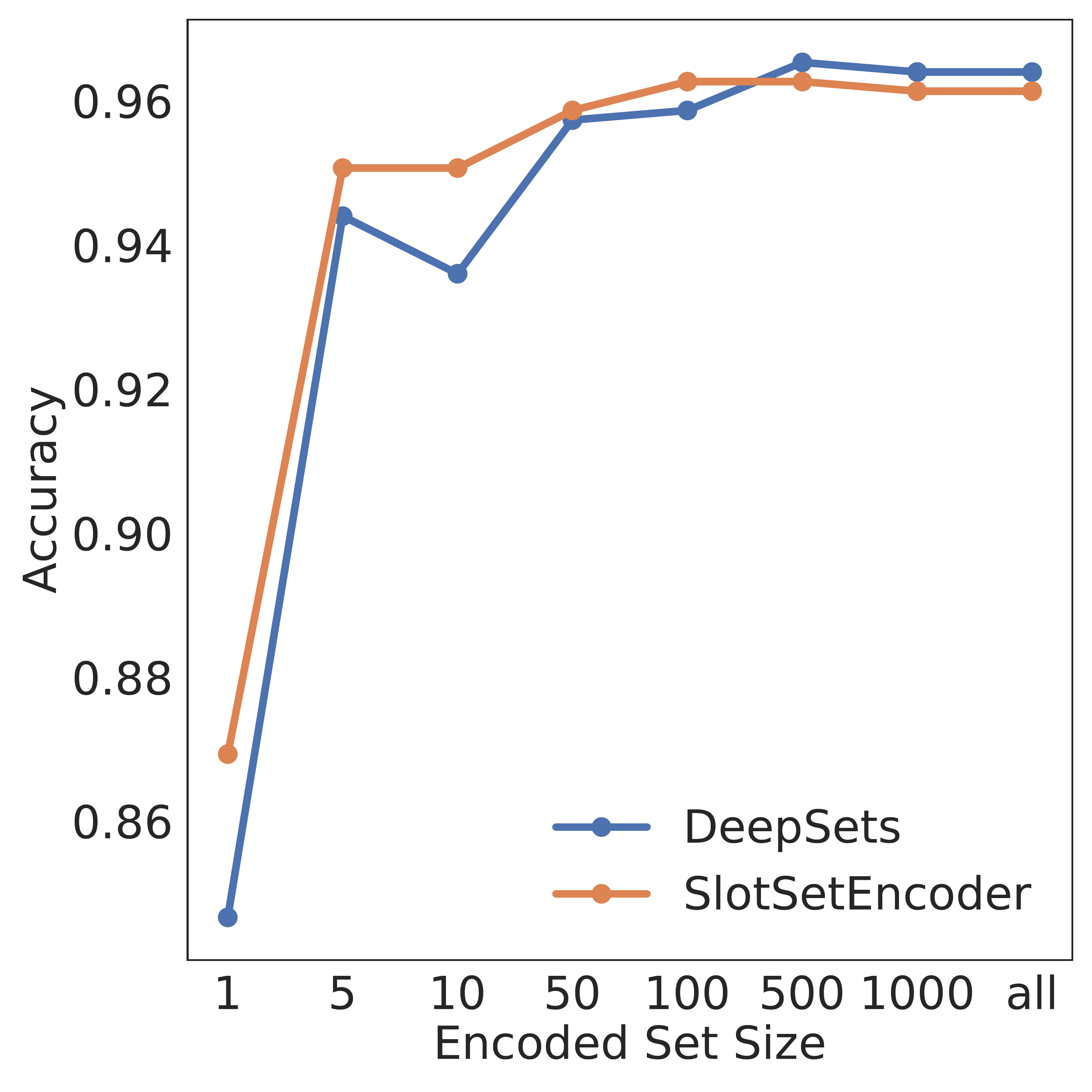}
		\caption{\small 5-way}
		\label{fig:5way}
	\end{subfigure}%
	\begin{subfigure}{.25\linewidth}
		\centering
		\includegraphics[width=\textwidth]{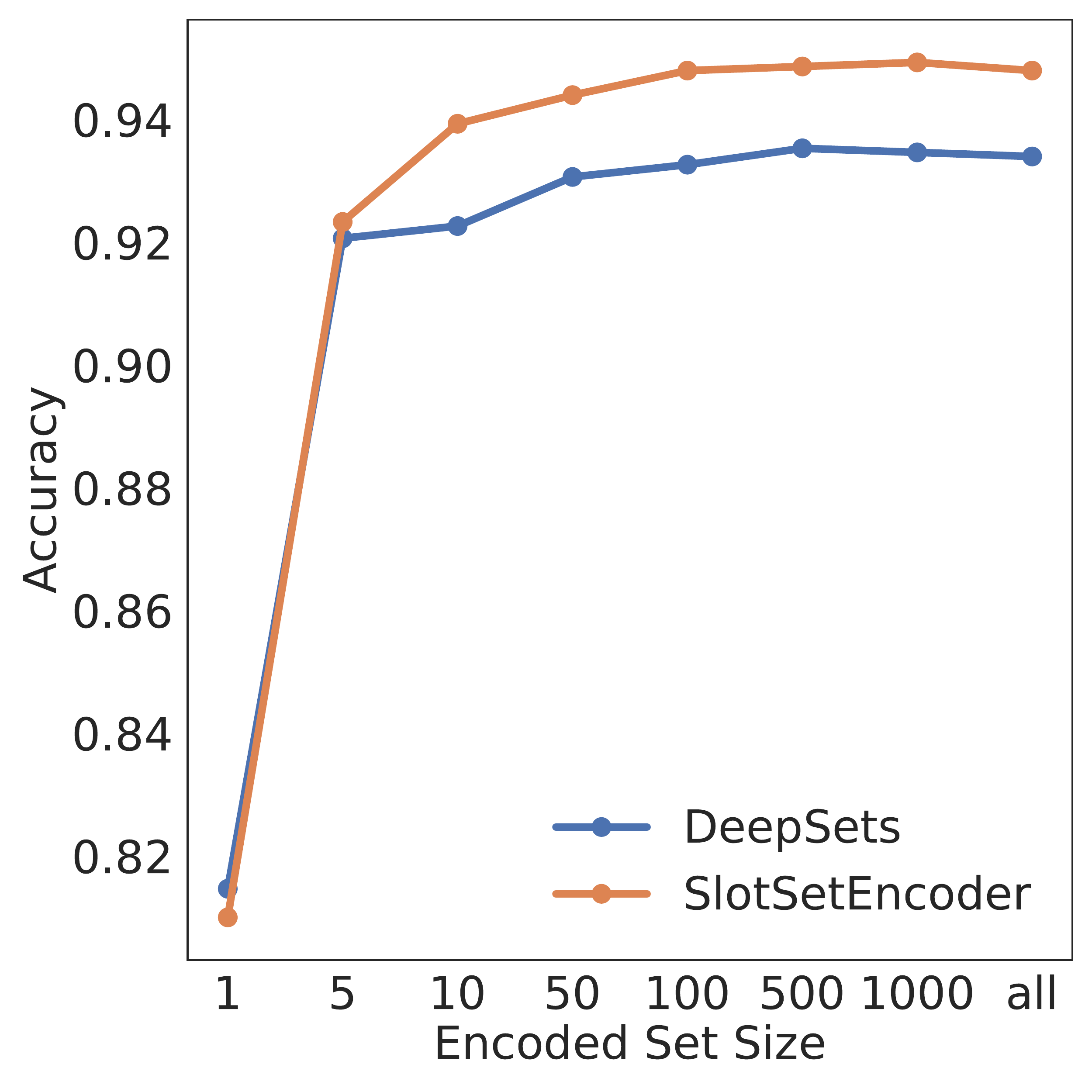}
		\caption{\small 10-way}
		\label{fig:10way}
	\end{subfigure}%
	\begin{subfigure}{.25\linewidth}
		\centering
		\includegraphics[width=\textwidth]{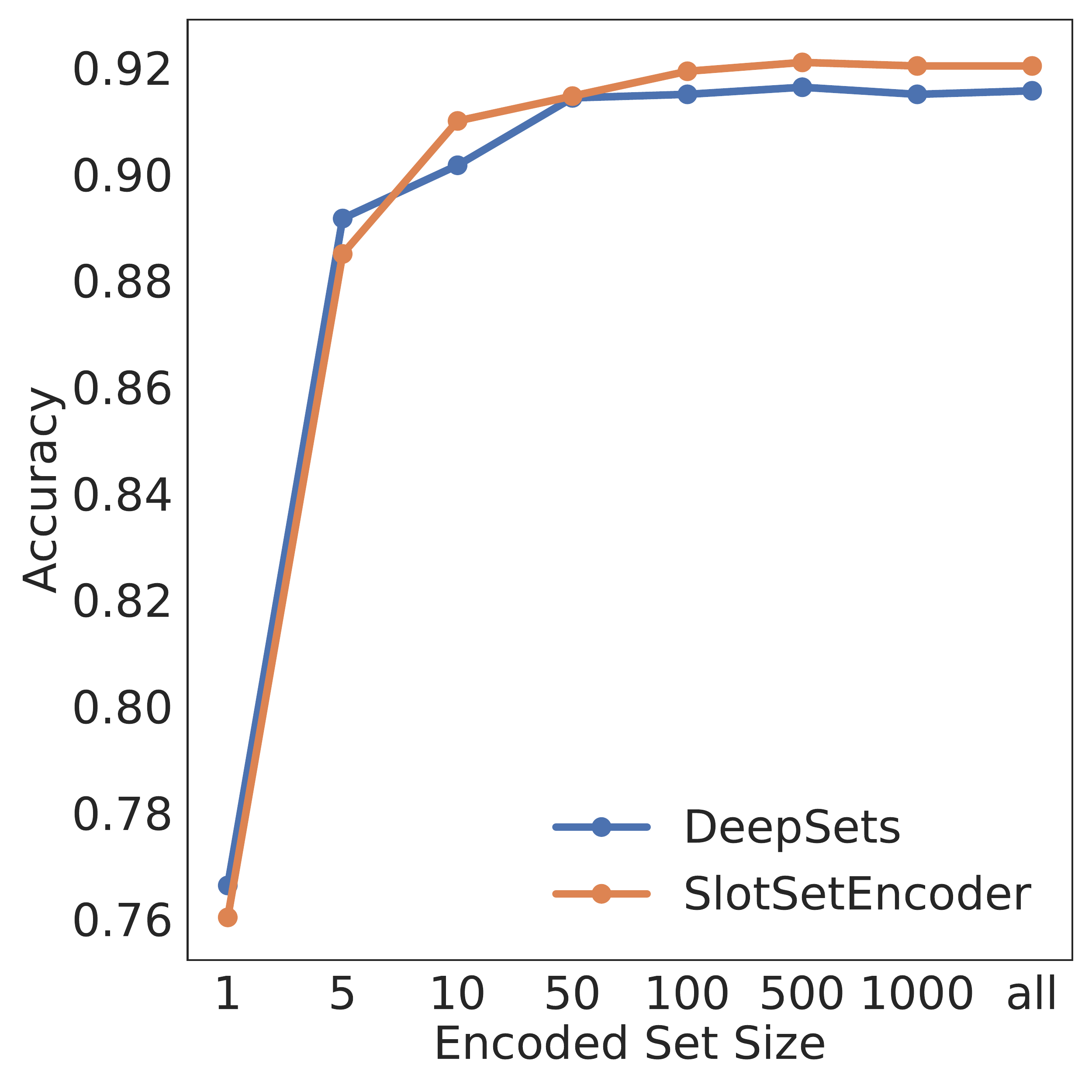}
		\caption{\small 20-way}
		\label{fig:20way}
	\end{subfigure}%
	\begin{subfigure}{.25\linewidth}
		\centering
		\includegraphics[width=\textwidth]{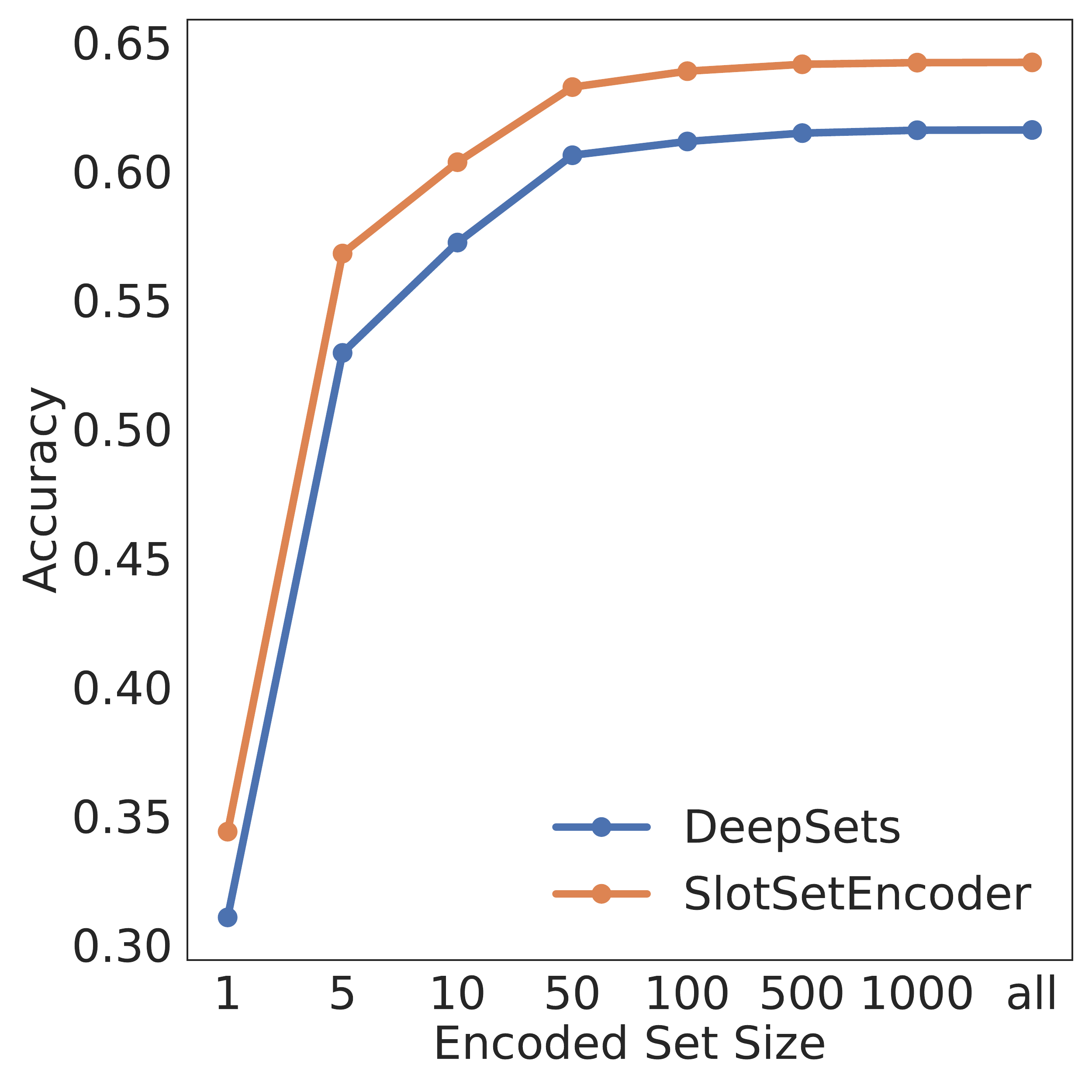}
		\caption{\small 1000-way}
		\label{fig:1000way}
	\end{subfigure}
    \caption{Accuracy vs. Encoded set sizes for different n-way classification problems on ImageNet. Our Slot Set Encoder performs best on n-way problems which differ from what was seen during training (trained on 20-way). 1-way provides a singleton set, and therefore is about equal for both models.}
    \label{fig:centroid_prediction_shots}
    \end{center}
\end{figure*}

%% file: sections/related.tex
\section{Related Work}
\paragraph{Set Encoding}\label{rel:setencoding}
In DeepSets, \citet{deepsets} show that neural architectures for set data are required to obey the permutation 
invariance/equivariance symmetry and can be formulated as the following sum-decomposition of two functions, $f(\mathbf{x}_n) =
\tilde{f}\big(\sum_{i=1}^{n}\phi(\mathbf{x}_i)\big)$ where $\phi$ is a feature extractor and $\tilde{f}$ is some non-linear 
activation function such as ReLU. Additionally, instead of the $\mathrm{sum}$ operator, $\mathrm{min, max}$ and $\mathrm{mean}$ 
also satisfy the imposed constraints. While there exist previous works \citep{causalsignals} on invariant pooling methods, 
DeepSets provided a universal model for constructing such neural networks. However, the architectures proposed 
in DeepSets are simple and in some cases lack expressivity.  Set Transformer \citep{settransformer} solves this issue by formulating a 
set compatible version of the Transformer \citep{transformer} which is capable of modeling pairwise interactions among the input set. Additionally, \citet{settransformer} reduce the complexity of Transformers from $\mathcal{O}(n^2)$ to $\mathcal{O}(mn)$
by introducing inducing points that serve as queries to reduce the size of the input set before self-attention. Other methods that tackle the set encoding problem include ~\citet{janossy} and ~\citet{fspool}. However all of these methods require processing all elements of the input set at once and hence cannot be used in the MBC setting.

\paragraph{Slot Attention}\label{rel:slotattention}
\citet{slotattention} introduced a slot attention mechanism to learn object-centric representations. Slots are variables 
which adapt and bind to objects (sets of features) to form non-stationary, input-dependent representations. The slot attention mechanism works by 
iteratively updating an initialized set of slots with a GRU \citep{gru} to adapt them to the current inputs. Through the adaptation process, the 
slots compete for sets of similar features 
which compose objects by specializing their weights to achieve a higher activation value in the softmax of the attention matrix. 
While the idea of slots provided an initial inspiration for our work, they were not designed for set encoding like DeepSets and break MBC for a couple of reasons:
1) The GRU iteratively updates the slots, which depends heavily on the set elements. 2) the Softmax normalization layer depends on all
elements in the set and 3) updates on hidden states in the model also break MBC.

%% file: sections/conclusion.tex
\section{Conclusion}\label{sec:conclusion}
In this work, we have identified a key limitation of set encoding methods when applied to large scale / streaming set encoding. Additionally, we identified
a key property, Mini-Batch Consistency, that is required to guarantee that set encoding methods
are amenable to mini-batch processing of set. We further detailed the formulation of an instance of 
an attentive set encoding mechanism dubbed Slot Set Encoder that respects the symmetries of permutation invariance/equivariance, 
is Mini-Batch Consistent and can be trained with a mini-batch approximation of the full set gradients. We demonstrated the utility
of SSE with extensive experiments and ablation studies and verified that SSE is capable of learning rich set representations while satisfying MBC.
An interesting direction  of future research would be to use more sophisticated methods to approximate the mini-batch training routine (in Section
\ref{sec:appminibatch}) with respect to the full set without having to consider all elements in the set.

\section{Acknowledgment}\label{sec:ack}
This work was supported by the Engineering Research Center Program through the National Research Foundation of Korea (NRF) funded by the Korean Government MSIT (NRF-2018R1A5A1059921), the Institute of Information \& communications Technology Planning \& Evaluation (IITP) grant funded by the Korea government (MSIT) (No.2020-0-00153), the National Research Foundation of Korea (NRF) funded by the Ministry of Education (NRF-2021R1F1A1061655), the Institute of Information \& communications Technology Planning \& Evaluation (IITP) grant funded by the Korea government(MSIT) (No.2021-0-02068, Artificial Intelligence Innovation Hub) and the Institute of Information \& communications Technology Planning \& Evaluation (IITP) grant funded by the Korea government(MSIT)  (No.2019-0-00075, Artificial Intelligence Graduate School Program(KAIST)).

%% file: sections/appendix.tex
\input{appendix/organization}
\input{appendix/slotsetencoder}
\input{appendix/propositions}
\input{appendix/image_reconstruction}
\input{appendix/pointcloud_classification}
\input{appendix/cluster_centroid}
\input{appendix/datasets}

%% file: appendix/organization.tex
\section{Organization}
The supplementary file is organized as follows: First in Section \ref{sec:slotsetencoder}, we provide a more detailed 
version of Algorithm 1 in the main paper. In Section \ref{sec:minibatchconsistency}, we provide proofs for Proposition 1 
and 2 and more details about the  mini-batch consistency of baseline models. Section \ref{sec:imagereconstruction} is on the Image Reconstruction experiments with the CelebA dataset where 
we provide details of the model architectures used. Additionally, we reproduce the plots in 
the main paper and include the deterministic slot initialization versions on the model together
with the plots for Slot Dimension and Slot Size mentioned in the main text. Section \ref{sec:pointcloud} provides 
the reference for the model architectures used for the Point Cloud classidication experiments
with the ModelNet40 dataset. Finally in Section \ref{sec:centroid_appendix}, we provide further details on the Cluster Centroid 
prediction problem.

\par Unless otherwise specified, all models were trained with the Adam optimizer with learning rate of $1e-3$ and weight-decay 
of $5e-4$. Additionally, all models are trained for 200 epochs with MultiStep learning rate scheduler 
at $0.2, 0.6, 0.8$ milestones of the full 200 epochs.

\subsection{Limitations \& Societal Impact}

\paragraph{Limitations.} As stated in section 2.5, our work considers a training time gradient which is an approximation to the gradient of a larger set which must be processed in batches. While effective, there could be opportunities for future research to approximate the gradient of the full set through episodic training, or other strategies which could improve performance.

\paragraph{Societal Impact.} Expressive set functions and have the potential for improving fairness in AI by creating more informative contributions from set elements. A permutation equivariant, sum-decomposable function such as Deep Sets grants equal weights to each latent feature, while attention used by both \citep{settransformer} and SSE could provide ways for underrepresented set elements to have more impact on the final prediction. Other than generalization error risk shared by all predictive models, we are unaware of any potential negative societal impacts of our work. 

%% file: appendix/slotsetencoder.tex
\section{Slot Set Encoder}\label{sec:slotsetencoder}
Below, we provide a version of the Slot Set Encoder Algorithm with finer detail than space allowed in the main text.
\input{algorithms/slotsetencoderapp}

%% file: algorithms/slotsetencoderapp.tex
\begin{algorithm}[H]
   \caption{Slot Set Encoder. $X = \{X_1, X_2, \ldots, X_p \}$ is the input set partitioned into $p$ chunks. $S\in \mathbb{R}^{K \times h}$ are the initialized slots and $g$ is the choice of aggregation function.}
   \label{alg:slotsetencoderapp}
  \begin{algorithmic}[1]
    \State {\bfseries Input:} $X = \{X_1, X_2, \ldots, X_p\}$, $S \in \mathbb{R}^{K \times h}, g$
    %\STATE {\bfseries Parameters:} $q, v, s$
    \State {\bfseries Output:} $\hat{S} \in \mathbb{R}^{K \times d}$
     
    \State \textbf{Initialize} $\hat{S}$ 
    \State $S = \verb|LayerNorm|(S)$
    \State $q = \verb|Linear|_q (S)$
    \For{$i=1$ {\bfseries to} $p$}
      \State $k = \verb|Linear|_k (X_i)$
      \State $v = \verb|Linear|_v (X_i)$
      \State $M = \frac{1}{\sqrt{\hat{d}}} * k \cdot q^{T}$
      \State $\verb|attn| = \verb|Sigmoid|(M) + 1e-8$
      \State $W = \verb|attn| / \verb|attn.sum(dim=2)|$
      \State $\hat{S_i} = W^{T} \cdot v$
      \State $\hat{S} = g(\hat{S}, \hat{S_i})$
    \EndFor
    \State \textbf{return} $\hat{S}$
  \end{algorithmic}
\end{algorithm}

%% file: appendix/propositions.tex
\section{Mini-Batch Consistency}\label{sec:minibatchconsistency}
We restate Propositions 1 \& 2 here.
\begin{property}[Mini-Batch Consistency]\label{prop:consistency_app}
  Let $X \in \mathbb{R}^{n \times d}$ be partitioned such that $X = X_1 \cup X_2 \cup \ldots \cup X_p$ and $f:\mathbb{R}^{n_i \times d} \mapsto
  \mathbb{R}^{d^\prime}$ be a set encoding function such that $f(X) = Z$. Given an aggregation function $g:\{Z_j \in \mathbb{R}^{d^\prime}\}_{j=1}^{p} \mapsto
  \mathbb{R}^{d^\prime}$, $g$ and $f$ are Mini-Batch Consistent if and only if
  \[g\big(f(X_1), \ldots, f(X_p)\big) = f(X)\]
\end{property}

\begin{proposition}\label{props:consistency_app}
  For a given input set $X \in \mathbb{R}^{n \times d}$ and slot initialization $S \in \mathbb{R}^{K \times d}$, the functions $f$ and $g$ as defined
  in Algorithm \ref{alg:slotsetencoder} are Mini-Batch Consistent for any partition of $X$ and hence satisfy Property \ref{prop:consistency_app}.
\end{proposition}

\begin{proposition}\label{props:perinvariance_app}
  Let $X \in \mathbb{R}^{n \times d}$ and $S \in \mathbb{R}^{K \times d}$ be an input set and slot initialization respectively.
  Additionally, let $\verb|SSEncoder|(X, S)$ be the output of Algorithm \ref{alg:slotsetencoder}, and $\pi_X \in \mathbb{R}^{n \times n}$ and
  $\pi_S \in \mathbb{R}^{K \times K}$ be arbitrary permutation matrices. Then,
  \[\verb|SSetEncoder|(\pi_X \cdot X, \pi_S \cdot S) = \pi_S \cdot \verb|SSEncoder|(X,S)\]
\end{proposition}

%\textcolor{blue}{To empirically check the validity of Propositions 1 \& 2 for the Slot Set Encoder, kindly check the file 
%models/slotsetencoder.py in the provided implementation at \url{https://github.com/AnonymousRepo1111/SSE }.}

\subsection{Proof of Proposition \ref{prop:consistency_app}}
We show that Algorithm~\ref{alg:slotsetencoder} satisfies Property~\ref{prop:consistency_app} by showing that each component of Algorithm~\ref{alg:slotsetencoder}
satisfies Property~\ref{prop:consistency_app}.

\begin{proof}
\textbf{Linear Layers.} Since all $\verb|Linear|$ layers in Algorithm~\ref{alg:slotsetencoder} are row-wise feed-forward neural network, they all satisfy 
Property~\ref{prop:consistency_app} since we can use concatenation to aggregate their outputs for any partition of the inputs. That is, all Linear 
layers in Algorithm~\ref{alg:slotsetencoder} satisfy Property~\ref{prop:consistency_app}.

\textbf{Dot Product.} \textbf{Equation 3 \& 4} compute the dot product over slots which involves the sum operation. Since the same slots are used for all 
elements in the input set, the dot product satisfies Property~\ref{prop:consistency_app} using similar arguments on Linear layers. The sum operation clearly 
satisfies Property~\ref{prop:consistency_app}.

\textbf{Choice of $g$.} Any choice of $g$ from the set $\verb|{sum, mean, max, min}|$ satisfies Property~\ref{prop:consistency_app}.

Since Algorithm~\ref{alg:slotsetencoder} is a composition of only the functions above, we conclude that Algorithm~\ref{alg:slotsetencoder} satisfies 
Property~\ref{prop:consistency_app}.
\end{proof}

\subsection{Proof of Proposition \ref{props:perinvariance_app}}
The proof is very closely tied to that provided in Appendix D of \citet{slotattention}. The definitions for permutation invariance and 
and equivariance are from Appendix D of \citet{slotattention}. We provide it here for completeness.

\begin{definition}[Permutation Invariance]
  A function $f:\mathbb{R}^{M \times D_1} \rightarrow \mathbb{R}^{M \times D_2}$ is permutation invariant if for any arbitrary permutation
  matrix $\pi \in \mathbb{R}^{M \times M}$ it holds that
  \[f(\pi x) = f(x)\]
\end{definition}

\begin{definition}[Permutation Equivariance]
  A function $f:\mathbb{R}^{M \times D_1} \rightarrow \mathbb{R}^{M \times D_2}$ is permutation equivariant if for any arbitrary permutation
  matrix $\pi \in \mathbb{R}^{M \times M}$ it holds that
  \[f(\pi x) = \pi f(x)\]
\end{definition}

We provide the proof for Proposition \ref{props:perinvariance_app} by considering each component of Algorithm \ref{alg:slotsetencoder}.

\begin{proof}
\textbf{Equation 3.} In  Algorithm \ref{alg:slotsetencoder}, we compute the following:
\begin{equation}
    \text{attn}_{i,j} := \sigma (M_{i,j})
\end{equation}
where $\sigma$ is the Sigmoid function. From the definition of the Sigmoid function, it follows that:
\begin{equation}
    \begin{split}
    \sigma (\pi_s \cdot \pi_k \cdot M_{i,j}) &= \text{Sigmoid}(\pi_s \cdot \pi_k \cdot M_{i,j}) \\
                     & = \frac{1}{1 + {\rm e}^{-(\pi_s \cdot \pi_k \cdot M)_{i,j}}} \\
                     & = \frac{1}{1 + {\rm e}^{-M_{\pi_{s(i)}, \pi_{k(j)}}}} \\
                     & = \sigma (M_{i,j})_{\pi_{s(i)}, \pi_{k(j)}}
    \end{split}
\end{equation}
That is, the Sigmoid function is permutation equivariant and $\pi_s$ and $\pi_k$ are permutation functions applied to $i$ and $j$ respectively. 

\textbf{Linear Layers.} $\verb|Linear|$ layers are independently applied to each of the inputs and the slots and hence are permutation equivariant. 

\textbf{Dot Product.} The dot product involves a sum over the feature dimension and hence also permutation equivariant. This makes Equation 3 \& 4, 
which make use of the dot product all permutation equivariant. 

\textbf{Choice of $g$.} Any choice of $g$ from the set $\verb|{sum, mean, max, min}|$ is permutation invariant. 

By combining all these operations, the Slot Set Encoder in Algorithm~\ref{alg:slotsetencoder} is permutation invariant with respect to the inputs 
and equivariant with with respect to a given set of slot initialization.
\end{proof}

\subsection{Mini-Batch Consistency of Baseline Models}\label{sec:baselines_mbc}

We provide an analysis of previously published set encoding models and whether or not their original formulation satisfies MBC. 

\begin{table}[!ht]
    \centering
    \begin{tabular}{lcl}
        \toprule
        Model & Pass/Fail & Reason \\
        \midrule
        Deepsets (original) & Fails &     Message Passing Layers \\
        Set Transformer & Fails &     Self-Attention  \\
        FSPool & Fails & Sorting \\
        \bottomrule
    \end{tabular}
    \caption{Pass/Fail test for Baselines satisfying Property~\ref{props:consistency_app}}
    \label{tab:baselines_mbc}
\end{table}

\paragraph{Message Passing Layers.} The original formulation of Deepsets~\citep{deepsets} utilizes simple message passing layers which use a set dependent normalization, where the considered set is only that which is present in the current batch. Specifically, each set in the current batch is normalized by $\phi(x) = f(x) - g(f(x))$, where $g \in \{max, mean\}$. Normalization based on batch statistics creates a dependency on the current batch and therefore causes the model to fail MBC. As it is trivial to make Deepsets satisfy MBC, our MBC version of Deepsets replaced these layers with traditional neural network layers, which satisfy MBC. 

\paragraph{Self-Attention.} Self attention used in the Set Transformer \citep{settransformer} allows for modelling complex interactions between set elements, but as depicted in the concept illustration in the main text, creates a dependency on the current batch while constructing the attention matrix $attn(X, X) = \sigma(q(X) * k(X)^\top)$, where $\sigma$ is the softmax function. Both the inner query-key multiplication and the softmax function are dependent on the current batch, thereby failing to satisfy MBC. 

%\paragraph{Sorting}

%% file: appendix/image_reconstruction.tex
\section{Image Reconstruction}\label{sec:imagereconstruction}
In our image reconstruction experiments, we use an encoder-decoder architecture with 
the encoder followed by a set encoding function. We provide details of the full 
model in Tables \ref{table:encoder} and \ref{table:decoder}. Additionally, we provide 
a the plots for the image reconstruction task with the version of the Slot Set Encoder 
that utilizes deterministic slot initialization together with the plots for the Slot 
Dimension and Slot Number experiments referenced in the Ablation section in the main paper 
in Figure \ref{fig:ablation_app}.

\input{figures/celeba/appendix}

\begin{table}[H]
\begin{center}
  \begin{tabular}{l c}
  \toprule
  Layers \\
  \midrule
    Linear(in\_features=5, out\_features=64) $\rightarrow$ ReLU \\
    Linear(in\_features=64, out\_features=64) $\rightarrow$ ReLU \\
    Linear(in\_features=64, out\_features=64)  \\
	SlotSetEncoder(K=1, h=64, d=64, d\_hat=64, g='sum', \_slots=Random)  \\
  \bottomrule
  \end{tabular}
\end{center}
\caption{\small Encoder for Image Reconstruction. \label{table:encoder}}
\end{table}

\begin{table}[H]
\begin{center}
  \begin{tabular}{l c}
  \toprule
  Layers \\
  \midrule
    Linear(in\_features=128, out\_features=128) $\rightarrow$ ReLU \\
    Linear(in\_features=128, out\_features=128) $\rightarrow$ ReLU \\
    Linear(in\_features=128, out\_features=128) $\rightarrow$ ReLU \\
    Linear(in\_features=128, out\_features=3)  \\
  \bottomrule
  \end{tabular}
\end{center}
\caption{\small Decoder for Image Reconstruction \label{table:decoder}}
\end{table}

%% file: figures/celeba/appendix.tex
\begin{figure}%[t]
    \begin{center}
	\begin{subfigure}{.25\linewidth}
		\centering
		\includegraphics[width=\textwidth]{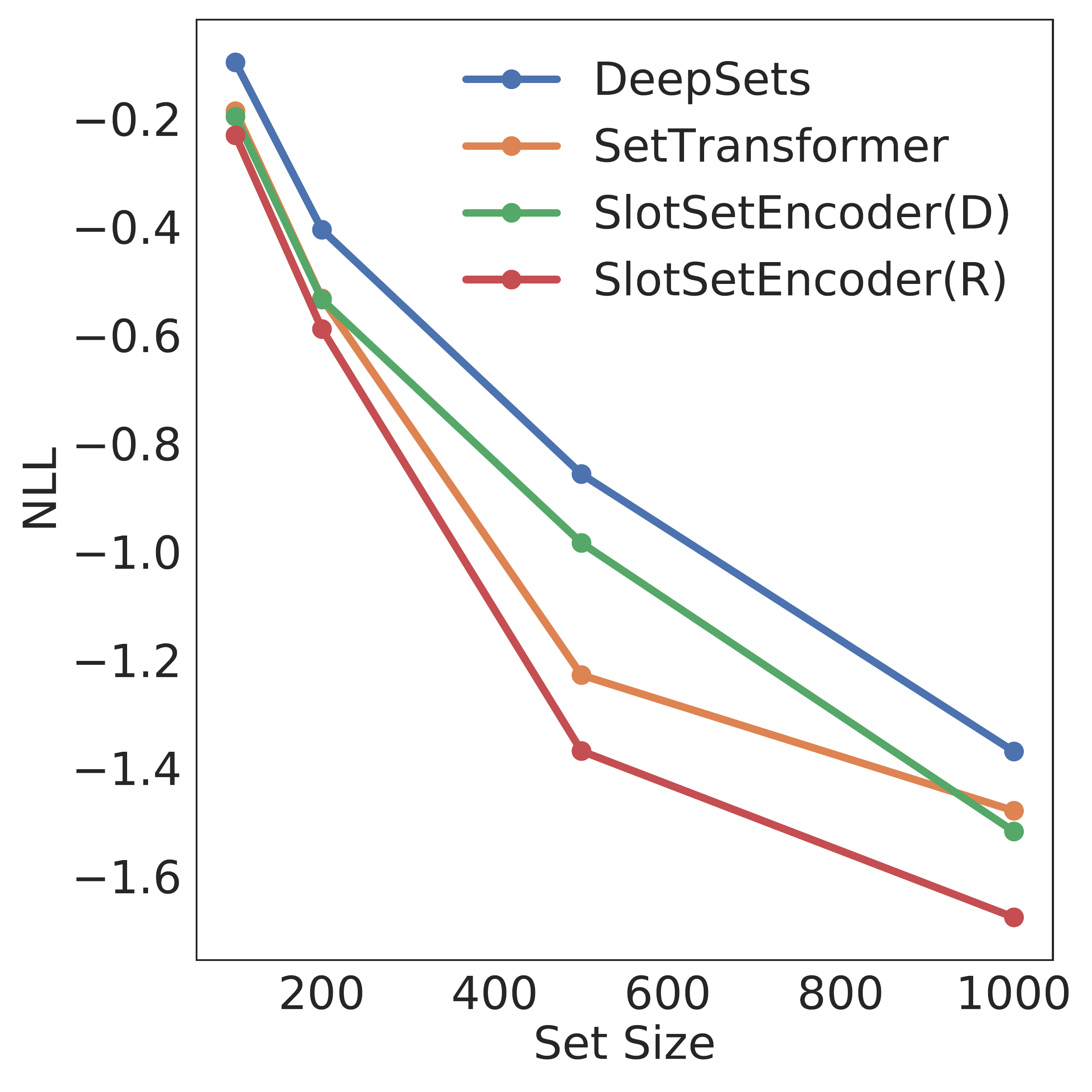}
		\caption{\small CelebA NLL }
		\label{fig:nll_celeba_app}
	\end{subfigure}%
	\begin{subfigure}{.25\linewidth}
		\centering
		\includegraphics[width=\textwidth]{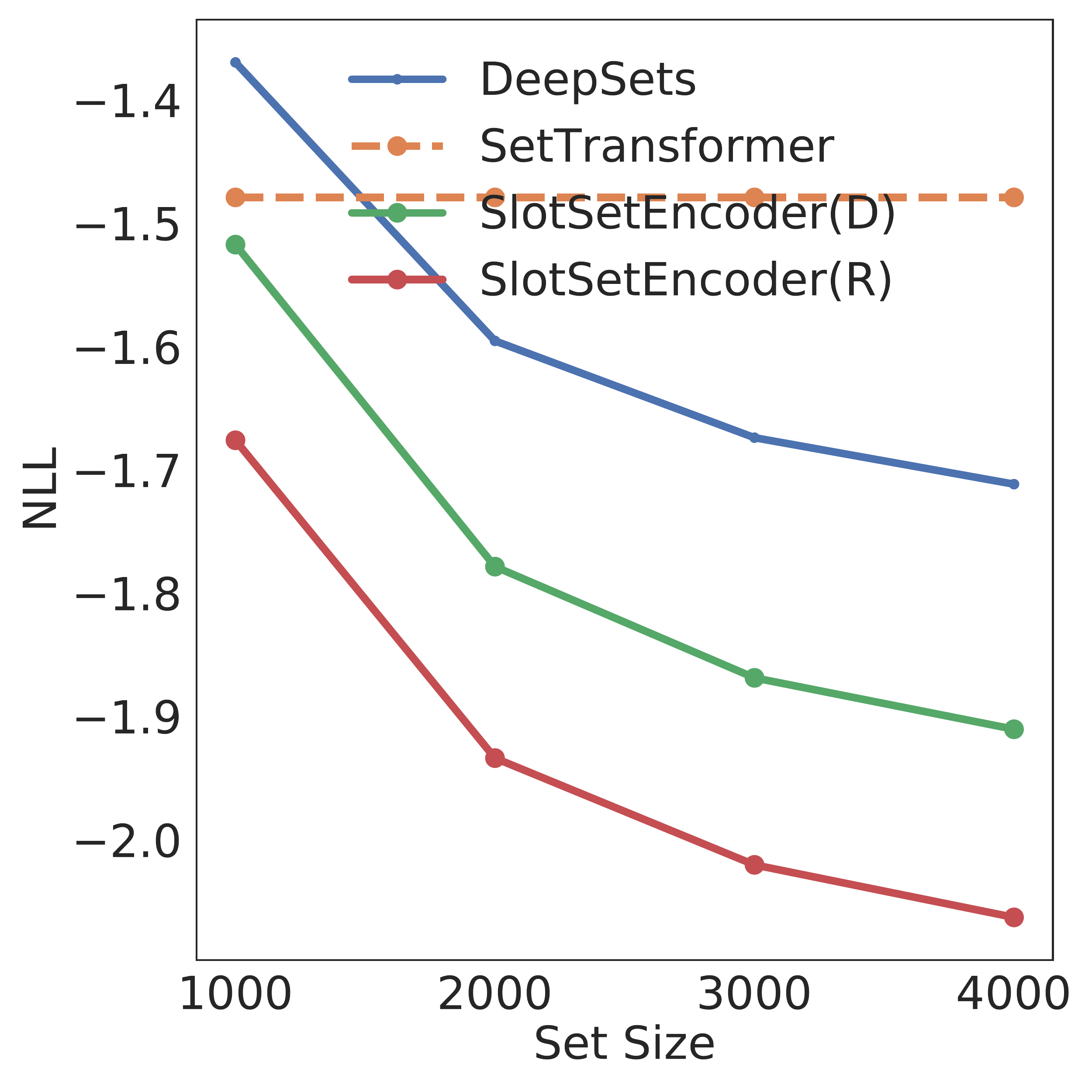}
		\caption{\small Mini-Batch Testing}
		\label{fig:generalization_app}
	\end{subfigure}%
	\begin{subfigure}{.25\linewidth}
		\centering
		\includegraphics[width=\textwidth]{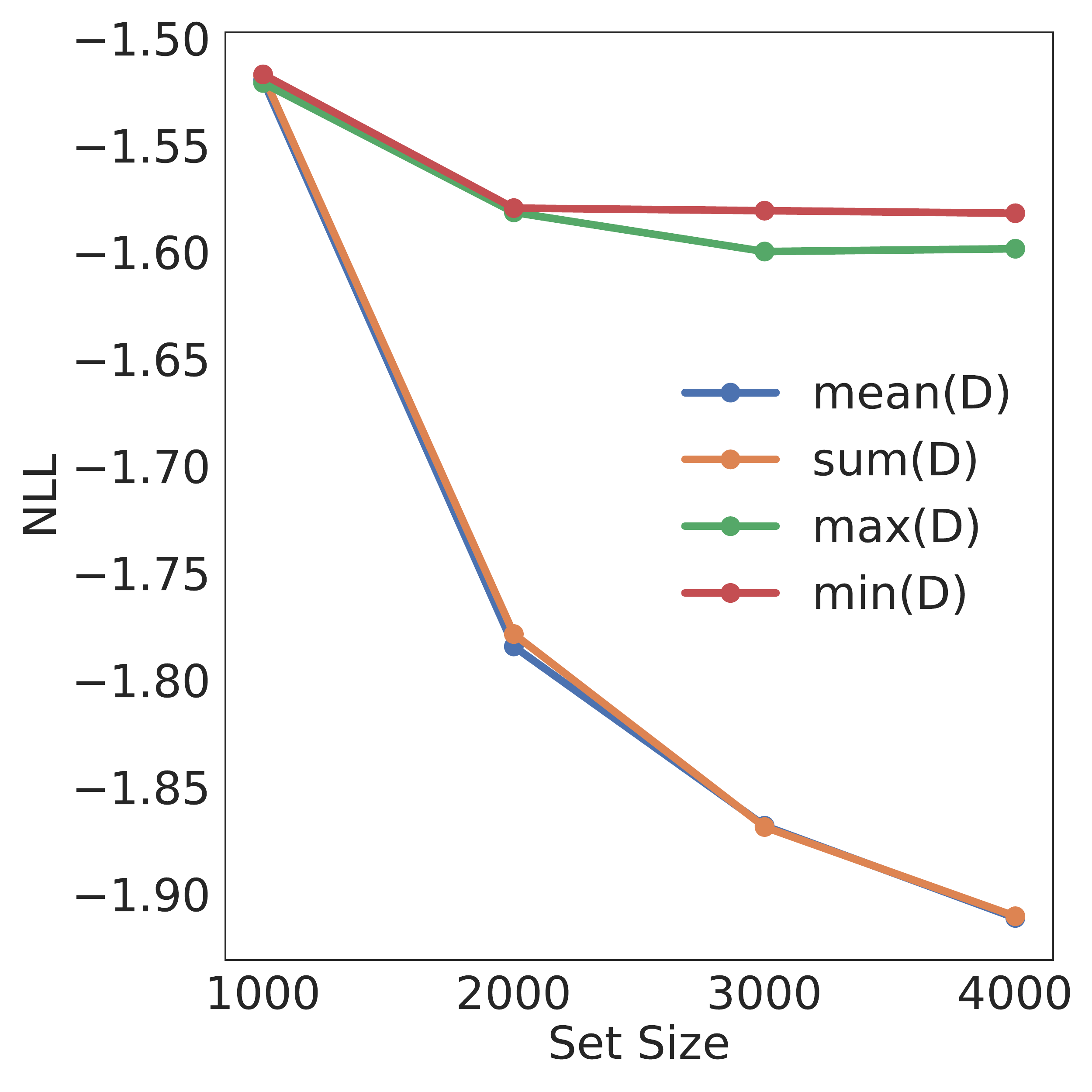}
		\caption{\small Choice of $g$ Deterministic}
		\label{fig:g_learned_app}
	\end{subfigure}%
	\begin{subfigure}{.25\linewidth}
		\centering
		\includegraphics[width=\textwidth]{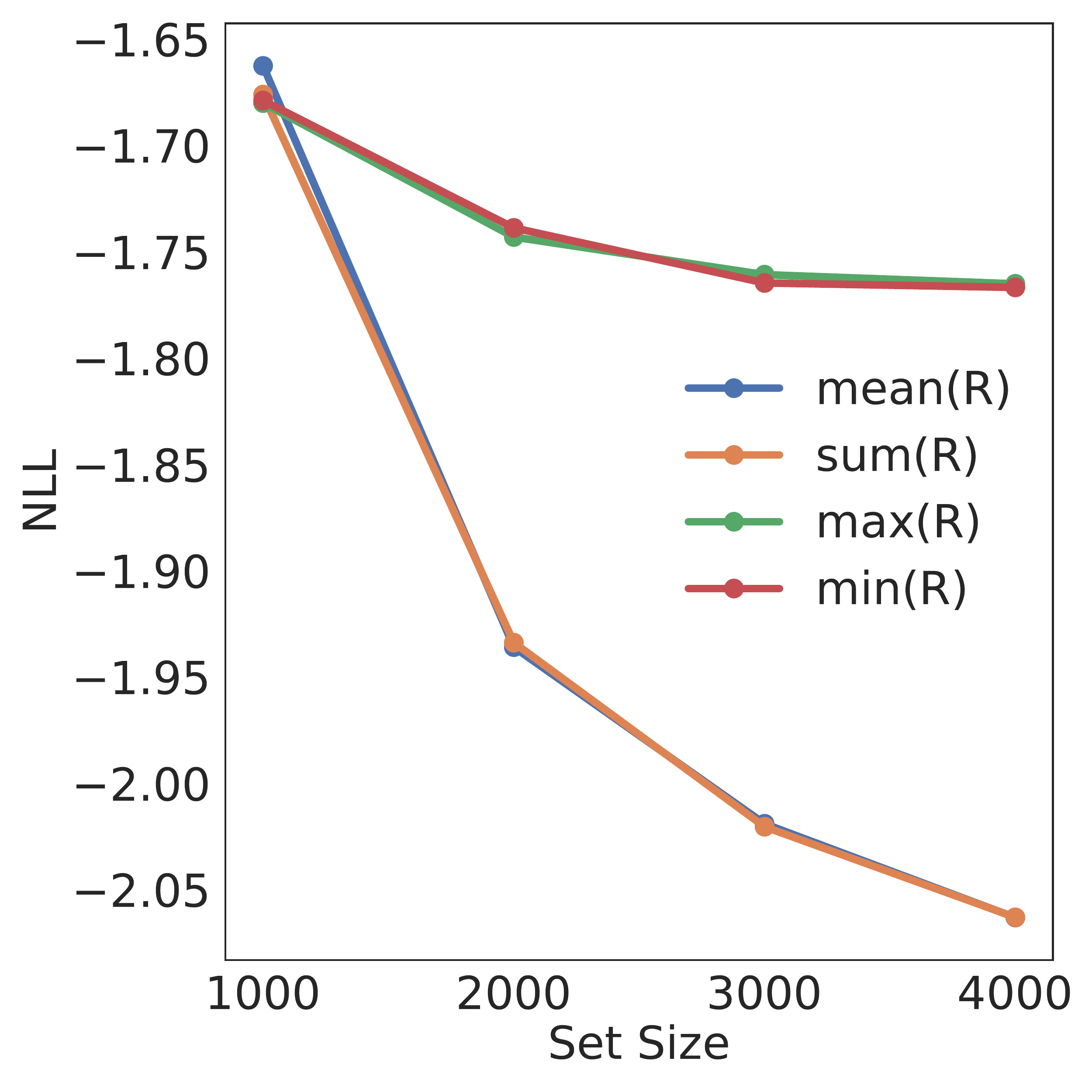}
		\caption{\small Choice of $g$ Random}
		\label{fig:g_random_app}
	\end{subfigure}
	\begin{subfigure}{.25\linewidth}
		\centering
		\includegraphics[width=\textwidth]{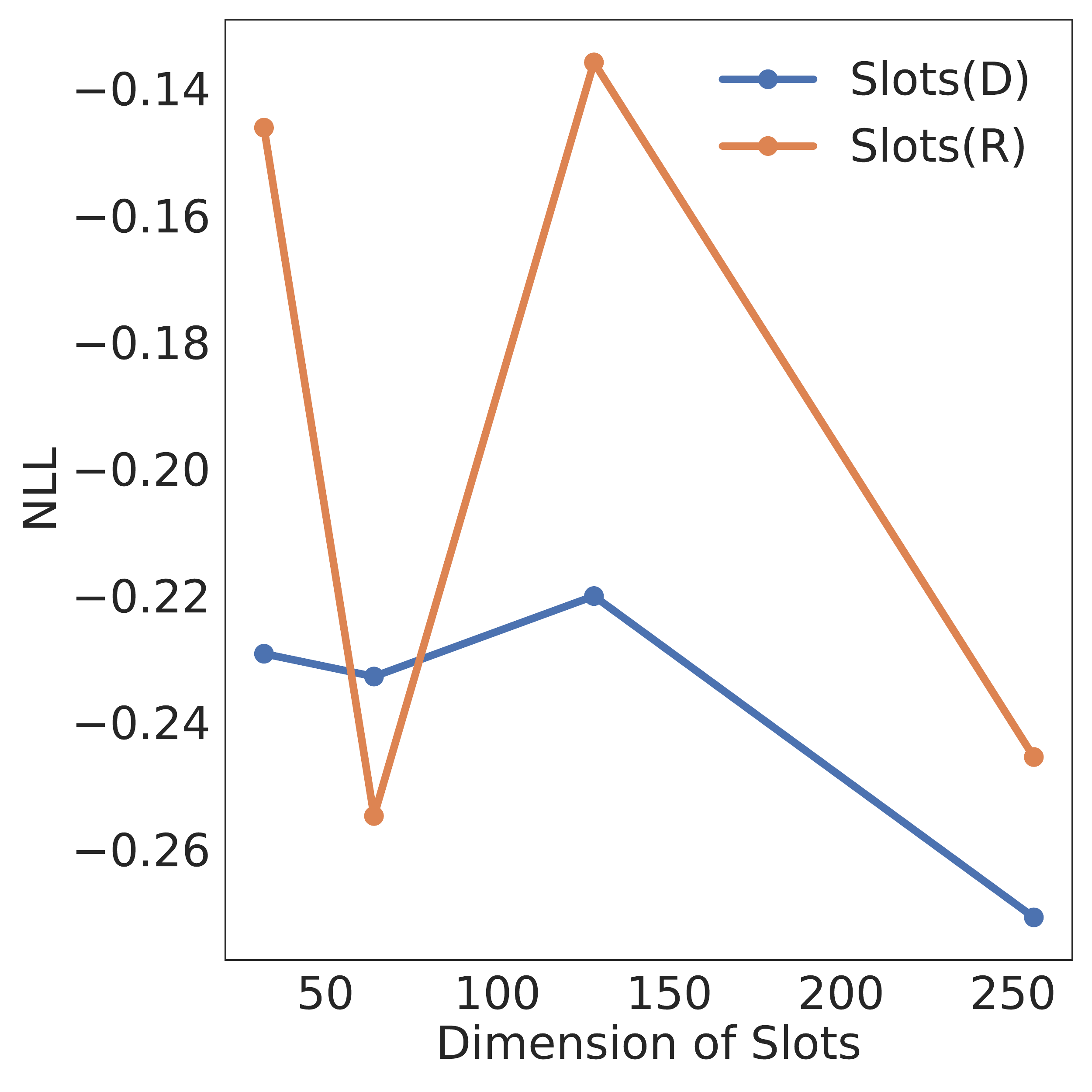}
		\caption{\small Slot Dimension}
		\label{fig:slot_dim_app}
	\end{subfigure}%
	\begin{subfigure}{.25\linewidth}
		\centering
		\includegraphics[width=\textwidth]{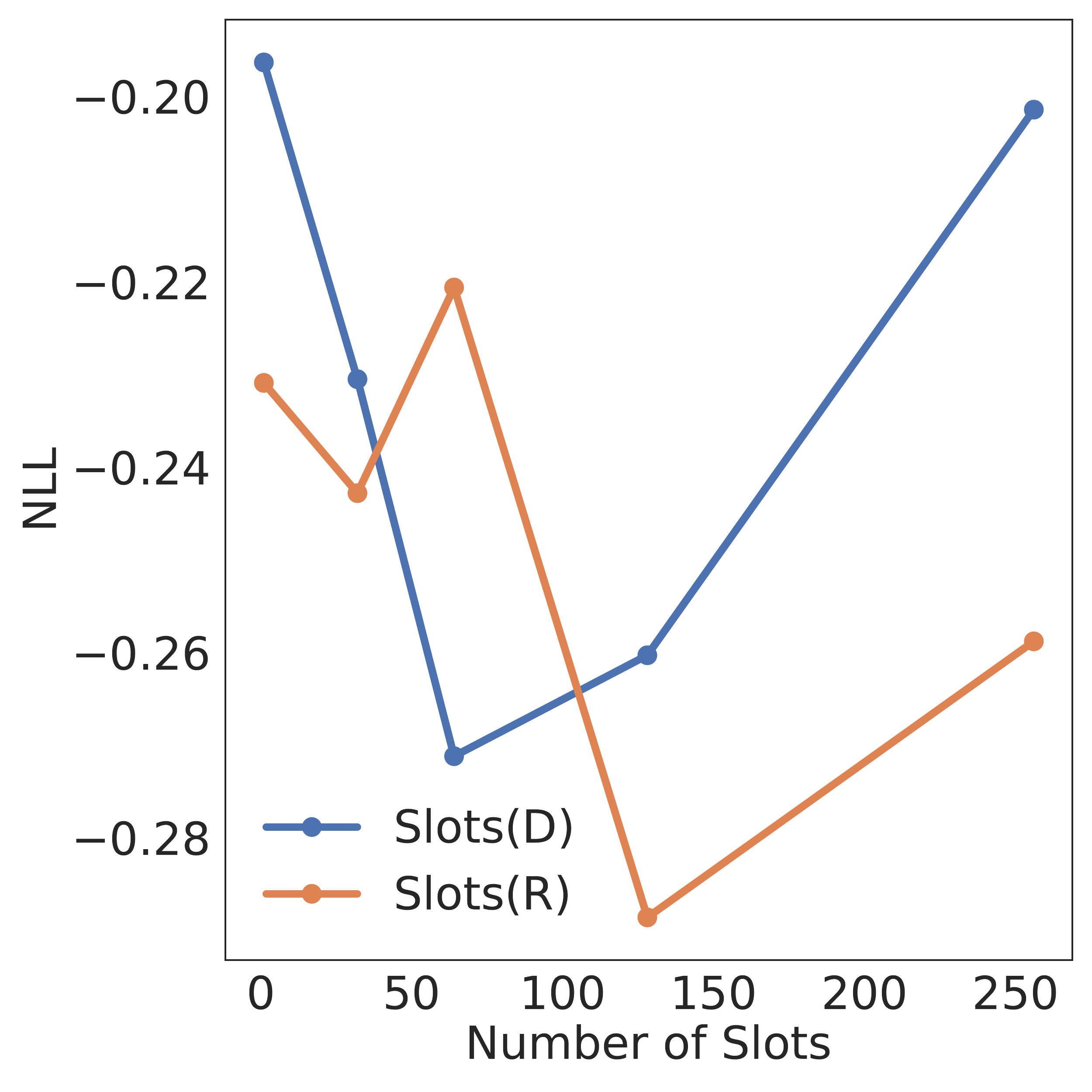}
		\caption{\small Number of Slots}
		\label{fig:num_slots_app}
	\end{subfigure}%
	\begin{subfigure}{.25\linewidth}
		\centering
		\includegraphics[width=\textwidth]{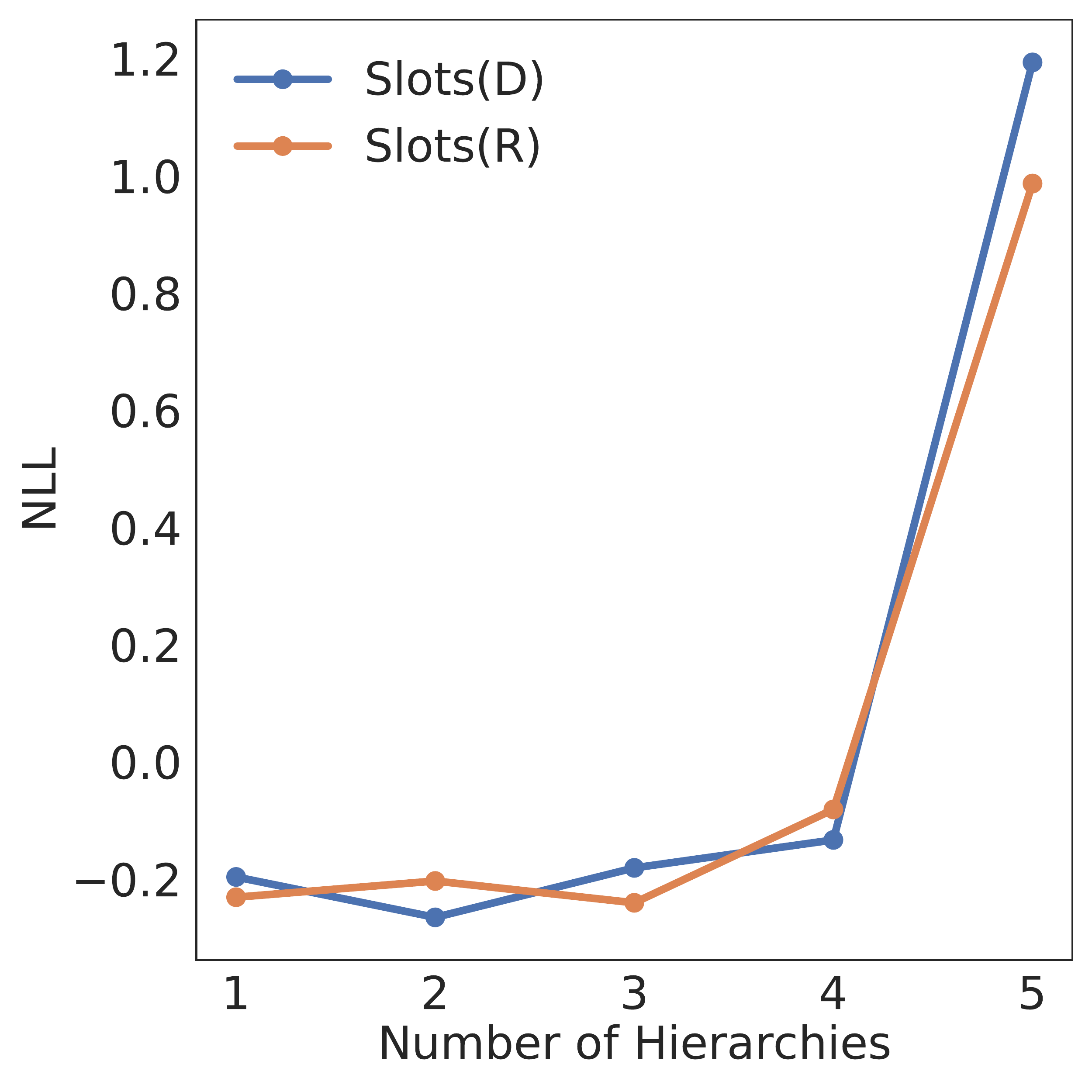}
		\caption{\small Number of Heirarchies}
		\label{fig:num_heir_app}
	\end{subfigure}%
	\begin{subfigure}{.25\linewidth}
		\centering
		\includegraphics[width=\textwidth]{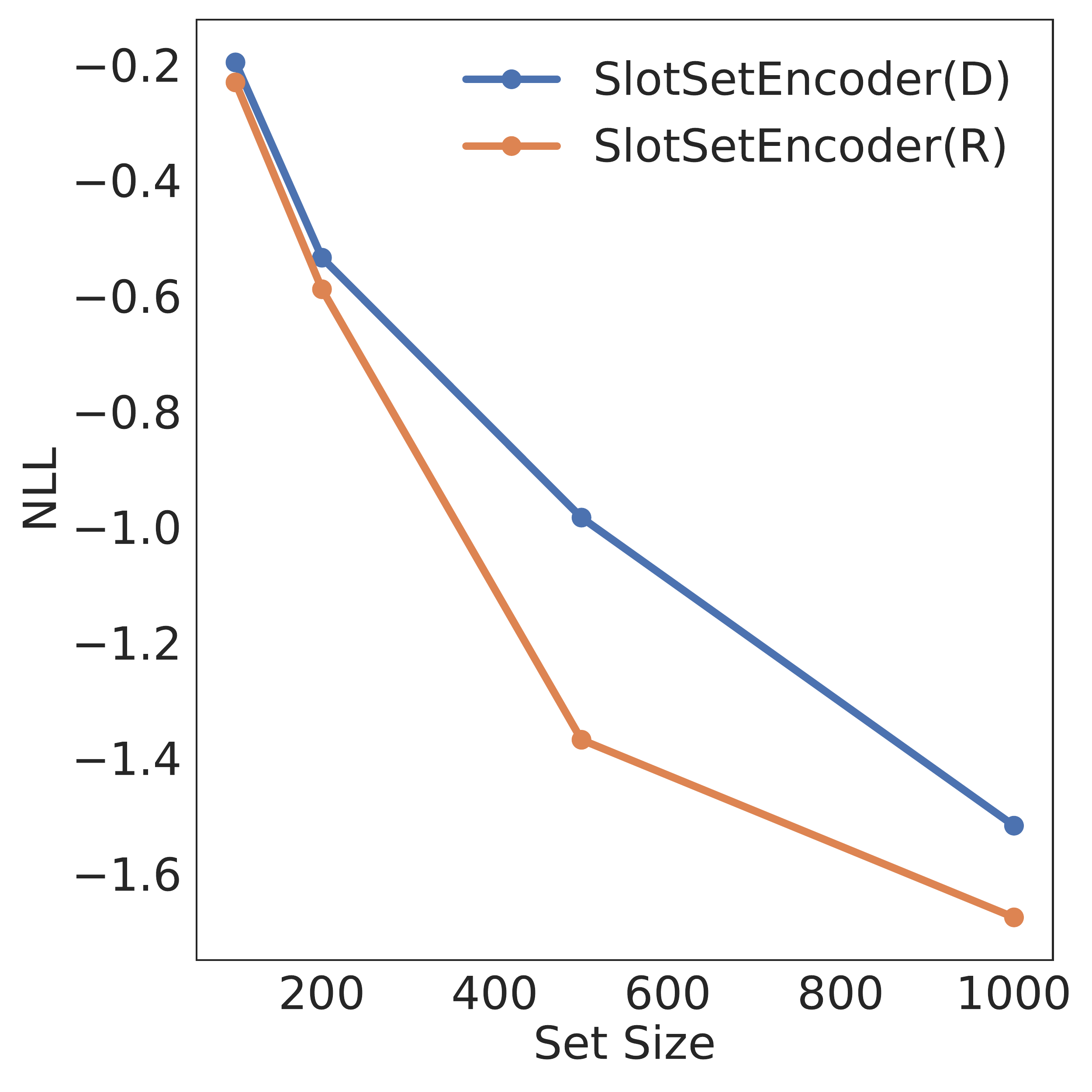}
		\caption{\small Random vs Deterministic}
		\label{fig:r_v_l_app}
	\end{subfigure}
	\caption{\small We provide the results in the main paper(Figures 2-6). Additionally, we add the result of the model with 
	deterministic slot initialization. \label{fig:ablation_app}}
	\end{center}
\end{figure}

%% file: appendix/pointcloud_classification.tex
\section{Point Cloud Classification}\label{sec:pointcloud}
For the ModelNet experiments on point cloud classification, we use the same architecture as \citet{deepsets} and replace 
the pooling layer with the Slot Set Encoder.

%% file: appendix/cluster_centroid.tex
\section{Cluster Centroid Architecture and Extra Results}\label{sec:centroid_appendix}
Below, we provide the exact architecture for the cluster centroid prediction task in section 3.3 for both the Deepsets model and our Mini-Batch Consistent Slot Set Encoder on MNIST/ImageNet. Both models were trained for a total of 50 epochs.

\clearpage
\subsection{MNIST}
\input{figures/centroid_prediction/centroid_prediction_line}
Each class in MNIST contains roughly 6000 training instances. During training, we randomly sample a set of 5 
classes with a support and query set, each consisting 10 instances from each class. Classwise support sets are encoded to a 128 dimensional vector (centroid). We follow the same training procedure as explained in section 3.3. Exact architectural dtails for the MNIST experiments can be found in Tables \ref{tbl:protonet_mbc_mnist} and \ref{tbl:protonet_deepsets_mnist}. For evaluation, we investigate the effects of encoding different set sizes (1-6000) on model performance. In Figure \ref{fig:centroid_prediction_tsne} we show the effect
of the encoded set size on the centroid location and observe that the centroid becomes more accurate as the set size increases. Precise centroids 
also lead to higher classification accuracy, highlighting the utility of encoding sets with high cardinality. This is further demonstrated in Figure
\ref{fig:centroid_prediciton_line} where one can see the Slot Set Encoder performs better on all set sizes compared to DeepSets.

\input{figures/centroid_prediction/centroid_prediction_tsne}
\input{figures/centroid_prediction/centroid_prediction_imagenet_tsne}

\begin{table}[H]
\begin{center}
  \begin{tabular}{l c}
  \midrule
  Layers \\
  \toprule
    Conv2d(1, 32) $\rightarrow$ BatchNorm $\rightarrow$ LeakyReLU $\rightarrow$ MaxPool(2) \\
    Conv2d(32, 64) $\rightarrow$ BatchNorm $\rightarrow$ LeakyReLU $\rightarrow$ MaxPool(2) \\
    Conv2d(64, 128) $\rightarrow$ BatchNorm $\rightarrow$ LeakyReLU $\rightarrow$ AvgPool \\
    SlotSetEncoder(K=32, dim=128) \\
    SlotSetEncoder(K=16, h=128, d=128, $\hat{d}$=128, g='mean') \\
  \bottomrule
  \end{tabular}
\end{center}
\caption{\small Architecture used for our Mini-Batch Consistent Slot Set Encoder on the MNIST centroid prediction task. \label{tbl:protonet_mbc_mnist}}
\end{table}

\begin{table}[H]
\begin{center}
  \begin{tabular}{l c}
  \toprule
  Layers \\
  \midrule
    Conv2d(1, 32) $\rightarrow$ BatchNorm $\rightarrow$ LeakyReLU $\rightarrow$ MaxPool(2) \\
    Conv2d(32, 64) $\rightarrow$ BatchNorm $\rightarrow$ LeakyReLU $\rightarrow$ MaxPool(2) \\
    Conv2d(64, 128) $\rightarrow$ BatchNorm $\rightarrow$ LeakyReLU $\rightarrow$ AvgPool \\
    DeepsetsMeanPooling(dim=setdim) \\
  \bottomrule
  \end{tabular}
\end{center}
  \caption{\small Architecture used for Deepsets on the MNIST centroid prediction task. \label{tbl:protonet_deepsets_mnist}}
\end{table}

\begin{table}[H]
\begin{center}
  \begin{tabular}{l c}
  \toprule
  Layers \\
  \midrule
    ResNet50(pretrained=True) \\
    SlotSetEncoder(K=128, dim=128) \\ 
  \bottomrule
  \end{tabular}
\end{center}
  \caption{\small Architecture used for SlotSetEncoder on the ImageNet centroid prediction task. \label{tbl:protonet_mbc_imagenet}}
\end{table}

\begin{table}[H]
\begin{center}
  \begin{tabular}{l c}
  \toprule
  Layers \\
  \midrule
    ResNet50(pretrained=True) \\
    Linear(128) $\rightarrow$ ReLU $\rightarrow$ Linear(128) \\ 
    DeepsetsMeanPooling(dim=setdim) \\
  \bottomrule
  \end{tabular}
\end{center}
  \caption{\small Architecture used for Deepsets on the ImageNet centroid prediction task. \label{tbl:protonet_deepsets_imagenet}}
\end{table}

%% file: figures/centroid_prediction/centroid_prediction_line.tex
\begin{wrapfigure}{r}{0.30\textwidth}
    \vspace{-0.6cm}
    \centering
    \includegraphics[width=0.30\textwidth]{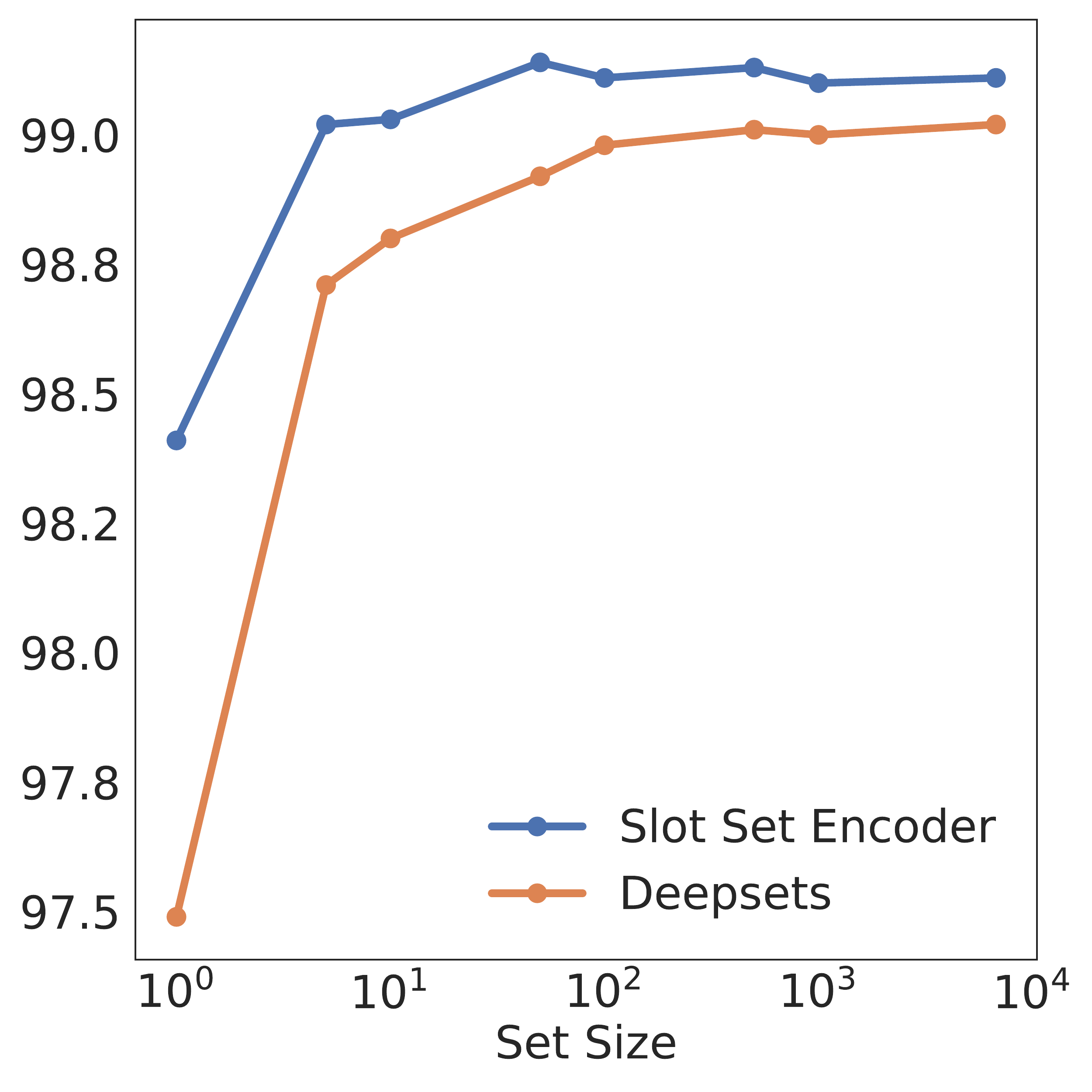}
    \caption{Accuracy vs. set size in the centroid prediction task. \label{fig:centroid_prediciton_line}}
    \vspace{-0.0cm}
\end{wrapfigure}

%% file: figures/centroid_prediction/centroid_prediction_tsne.tex
\begin{figure}[H]
    \centering
    \includegraphics[width=\textwidth]{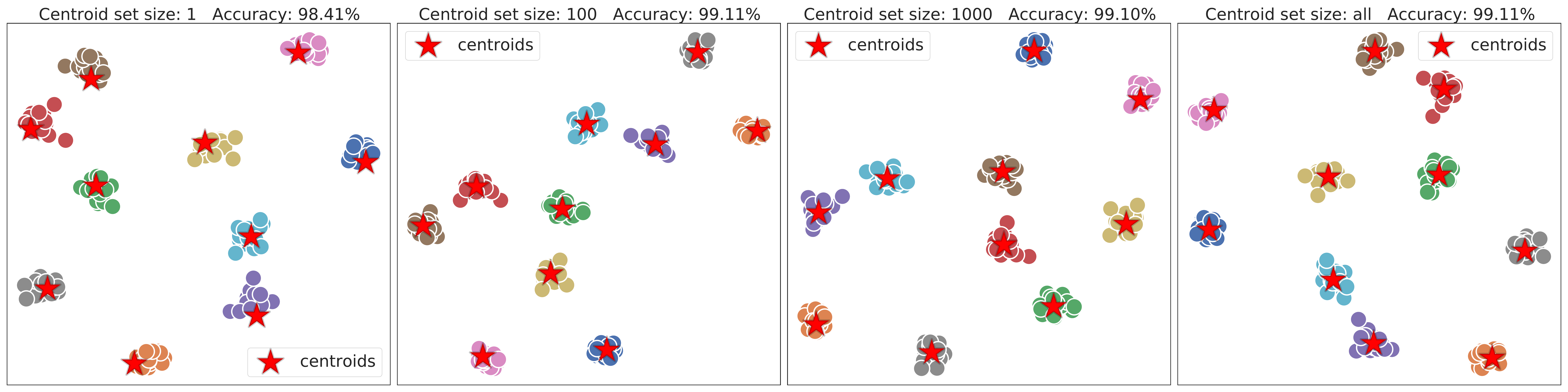}
    \caption{TSNE embeddings of centroids (stars) and classes (circles) in prototypical network \citep{protonet} classification. (left) a single instance, and (right) the entire training set for each class is encoded into a class centroid which is then used to classify the test set instances via Euclidian distance. As the set size used to predict the centroid increases, centroid location and classification accuracy both rise.}
    \label{fig:centroid_prediction_tsne}
\end{figure}

%% file: figures/centroid_prediction/centroid_prediction_imagenet_tsne.tex
\begin{figure}[H]
    \centering
    \includegraphics[width=\textwidth]{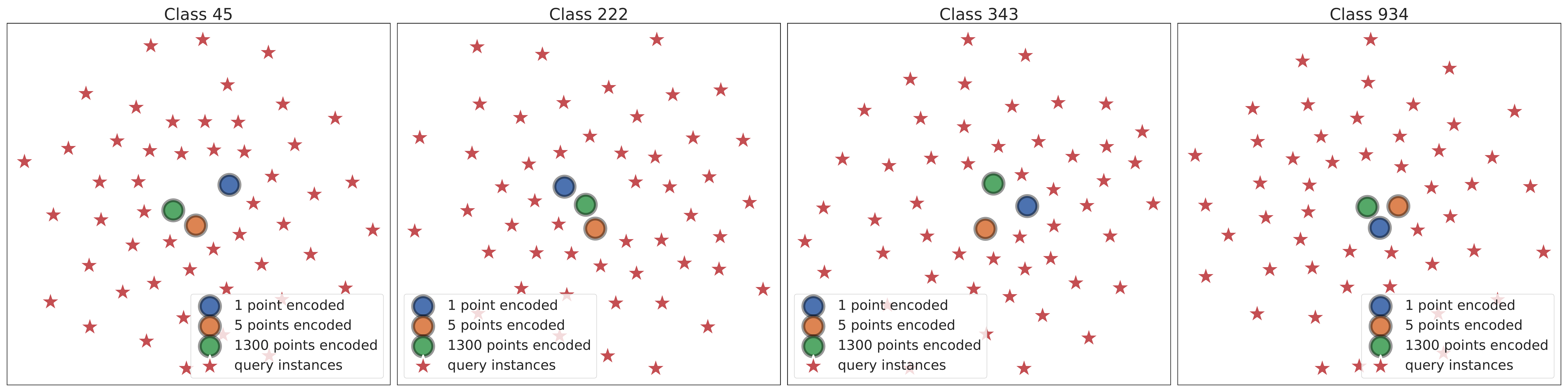}
    \caption{TSNE embeddings of centroids and query set in prototypical network \citep{protonet} classification. As the number of query instances embedded goes up, the centroid becomes more accurate. (See Figures 5 and 6 in the main text)
    \label{fig:centroid_prediction_imagenet_tsne}}
\end{figure}

%% file: appendix/datasets.tex
\section{Datasets}
We evaluate our model on the ImageNet \citep{imagenet}, CelebA \citep{celeba}, MNIST \citep{mnist} and ModelNet40 \citep{modelnet} datasets. MNIST consists 
of 60,000 training images of the handwritten digits 0-9 along with 10,000 test instances. CelebA consists of more than 200000 face images of celebrities with a wide
variety of pose variations and backgrounds.  ModelNet40 is as 3D CAD dataset with vertices for rendering common object with 40 classes. Point clouds
can be constructed from these vertices by sampling points from the vertices. The dataset consists of 9843 training instances and 2468 test instances.

%% file: neurips_2021.bbl
\begin{thebibliography}{15}
\providecommand{\natexlab}[1]{#1}
\providecommand{\url}[1]{\texttt{#1}}
\expandafter\ifx\csname urlstyle\endcsname\relax
  \providecommand{\doi}[1]{doi: #1}\else
  \providecommand{\doi}{doi: \begingroup \urlstyle{rm}\Url}\fi

\bibitem[Cho et~al.(2014)Cho, van Merrienboer, Gulcehre, Bahdanau, Bougares,
  Schwenk, and Bengio]{gru}
K.~Cho, B.~van Merrienboer, C.~Gulcehre, D.~Bahdanau, F.~Bougares, H.~Schwenk,
  and Y.~Bengio.
\newblock Learning phrase representations using rnn encoder-decoder for
  statistical machine translation, 2014.

\bibitem[Deng et~al.(2009)Deng, Dong, Socher, Li, Li, and Fei-Fei]{imagenet}
J.~Deng, W.~Dong, R.~Socher, L.-J. Li, K.~Li, and L.~Fei-Fei.
\newblock Imagenet: A large-scale hierarchical image database.
\newblock In \emph{2009 IEEE conference on computer vision and pattern
  recognition}, pages 248--255. Ieee, 2009.

\bibitem[Garnelo et~al.(2018)Garnelo, Rosenbaum, Maddison, Ramalho, Saxton,
  Shanahan, Teh, Rezende, and Eslami]{cnp}
M.~Garnelo, D.~Rosenbaum, C.~J. Maddison, T.~Ramalho, D.~Saxton, M.~Shanahan,
  Y.~W. Teh, D.~J. Rezende, and S.~Eslami.
\newblock Conditional neural processes.
\newblock \emph{arXiv preprint arXiv:1807.01613}, 2018.

\bibitem[Kim et~al.(2019)Kim, Mnih, Schwarz, Garnelo, Eslami, Rosenbaum,
  Vinyals, and Teh]{anp}
H.~Kim, A.~Mnih, J.~Schwarz, M.~Garnelo, A.~Eslami, D.~Rosenbaum, O.~Vinyals,
  and Y.~W. Teh.
\newblock Attentive neural processes.
\newblock \emph{arXiv preprint arXiv:1901.05761}, 2019.

\bibitem[LeCun et~al.(1998)LeCun, Bottou, Bengio, and Haffner]{mnist}
Y.~LeCun, L.~Bottou, Y.~Bengio, and P.~Haffner.
\newblock Gradient-based learning applied to document recognition.
\newblock \emph{Proceedings of the IEEE}, 86\penalty0 (11):\penalty0
  2278--2324, 1998.

\bibitem[Lee et~al.(2019)Lee, Lee, Kim, Kosiorek, Choi, and
  Teh]{settransformer}
J.~Lee, Y.~Lee, J.~Kim, A.~Kosiorek, S.~Choi, and Y.~W. Teh.
\newblock Set transformer: A framework for attention-based
  permutation-invariant neural networks.
\newblock In \emph{International Conference on Machine Learning}, pages
  3744--3753. PMLR, 2019.

\bibitem[Liu et~al.(2015)Liu, Luo, Wang, and Tang]{celeba}
Z.~Liu, P.~Luo, X.~Wang, and X.~Tang.
\newblock Deep learning face attributes in the wild.
\newblock In \emph{Proceedings of International Conference on Computer Vision
  (ICCV)}, December 2015.

\bibitem[Locatello et~al.(2020)Locatello, Weissenborn, Unterthiner, Mahendran,
  Heigold, Uszkoreit, Dosovitskiy, and Kipf]{slotattention}
F.~Locatello, D.~Weissenborn, T.~Unterthiner, A.~Mahendran, G.~Heigold,
  J.~Uszkoreit, A.~Dosovitskiy, and T.~Kipf.
\newblock Object-centric learning with slot attention.
\newblock \emph{arXiv preprint arXiv:2006.15055}, 2020.

\bibitem[Lopez-Paz et~al.(2017)Lopez-Paz, Nishihara, Chintala, Scholkopf, and
  Bottou]{causalsignals}
D.~Lopez-Paz, R.~Nishihara, S.~Chintala, B.~Scholkopf, and L.~Bottou.
\newblock Discovering causal signals in images.
\newblock In \emph{Proceedings of the IEEE Conference on Computer Vision and
  Pattern Recognition}, pages 6979--6987, 2017.

\bibitem[Murphy et~al.(2018)Murphy, Srinivasan, Rao, and Ribeiro]{janossy}
R.~L. Murphy, B.~Srinivasan, V.~Rao, and B.~Ribeiro.
\newblock Janossy pooling: Learning deep permutation-invariant functions for
  variable-size inputs.
\newblock \emph{arXiv preprint arXiv:1811.01900}, 2018.

\bibitem[Snell et~al.(2017)Snell, Swersky, and Zemel]{protonet}
J.~Snell, K.~Swersky, and R.~S. Zemel.
\newblock Prototypical networks for few-shot learning.
\newblock \emph{arXiv preprint arXiv:1703.05175}, 2017.

\bibitem[Vaswani et~al.(2017)Vaswani, Shazeer, Parmar, Uszkoreit, Jones, Gomez,
  Kaiser, and Polosukhin]{transformer}
A.~Vaswani, N.~Shazeer, N.~Parmar, J.~Uszkoreit, L.~Jones, A.~N. Gomez,
  {\L}.~Kaiser, and I.~Polosukhin.
\newblock Attention is all you need.
\newblock \emph{Advances in neural information processing systems},
  30:\penalty0 5998--6008, 2017.

\bibitem[Wu et~al.(2015)Wu, Song, Khosla, Yu, Zhang, Tang, and Xiao]{modelnet}
Z.~Wu, S.~Song, A.~Khosla, F.~Yu, L.~Zhang, X.~Tang, and J.~Xiao.
\newblock 3d shapenets: A deep representation for volumetric shapes.
\newblock In \emph{Proceedings of the IEEE conference on computer vision and
  pattern recognition}, pages 1912--1920, 2015.

\bibitem[Zaheer et~al.(2017)Zaheer, Kottur, Ravanbakhsh, Poczos, Salakhutdinov,
  and Smola]{deepsets}
M.~Zaheer, S.~Kottur, S.~Ravanbakhsh, B.~Poczos, R.~R. Salakhutdinov, and A.~J.
  Smola.
\newblock Deep sets.
\newblock In \emph{Advances in neural information processing systems}, pages
  3391--3401, 2017.

\bibitem[Zhang et~al.(2019)Zhang, Hare, and Pr{\"u}gel-Bennett]{fspool}
Y.~Zhang, J.~Hare, and A.~Pr{\"u}gel-Bennett.
\newblock Fspool: Learning set representations with featurewise sort pooling.
\newblock \emph{arXiv preprint arXiv:1906.02795}, 2019.

\end{thebibliography}
